\documentclass[journal]{IEEEtran}
\usepackage{verbatim}

\usepackage{cite}

\ifCLASSINFOpdf
   \usepackage[pdftex]{graphicx}
   \graphicspath{{./images/}}
\else
\fi
\usepackage[cmex10]{amsmath}
\usepackage{amssymb}
\usepackage{latexsym}
\usepackage{psfrag}
\usepackage{epsfig}
\usepackage{graphicx}
\usepackage{epstopdf}
\usepackage{multicol}
\usepackage{arydshln}
\usepackage{color}
\usepackage{tikz}
\usetikzlibrary{shapes,arrows}
\usepackage{verbatim}
\usepackage{subcaption}
\usepackage{caption}

\newtheorem{theorem}{Theorem}[section]

\newtheorem{proposition}[theorem]{Proposition}

\newtheorem{definition}[theorem]{Definition}
\newtheorem{remark}[theorem]{Remark}

\newcommand{\matt}[1]{\begin{bmatrix}#1\end{bmatrix}}

\newcommand{\R}{{\rm I\!R}}
\newcommand{\dfb}{\stackrel{\Delta}{=}}

\usepackage{url}

\begin{document}
%
\title{Controlling  rigid formations of mobile agents under inconsistent measurements}
%
%
%
\author{Hector~Garcia de Marina,~\IEEEmembership{Student Member,~IEEE,}
        Ming~Cao,~\IEEEmembership{Member,~IEEE,}
	and~Bayu~Jayawardhana,~\IEEEmembership{Senior Member,~IEEE}
\thanks{The authors are with the Institute of Technology, Engineering and Man-
agement, University of Groningen, 9747 AG Groningen, The Netherlands. (e-mail: \{h.j.de.marina, m.cao, b.jayawardhana\}.rug.nl). This work was supported by the the EU INTERREG program under the auspices of the SMARTBOT project and the work of Cao was also supported by the European Research Council
(ERC-StG-307207).}}


%
%

\markboth{IEEE TRANSACTIONS ON ROBOTICS}%
{}
%



\maketitle

\begin{abstract}
Despite the great success of using gradient-based controllers to stabilize rigid formations of autonomous agents in the past years, surprising yet intriguing undesirable collective motions have been reported recently when inconsistent measurements are used in the agents' local controllers. To make the existing gradient control robust against such measurement inconsistency, we exploit local estimators following the well known internal model principle for robust output regulation control. The new estimator-based gradient control is still distributed in nature and can be constructed systematically even when the number of agents in a rigid formation grows. We prove rigorously that the proposed control is able to guarantee exponential convergence and then demonstrate through robotic experiments and computer simulations that the reported inconsistency-induced orbits of collective movements are effectively eliminated.
\end{abstract}

\begin{IEEEkeywords}
	Formation Control, Distributed Control, Distributed Calibration
\end{IEEEkeywords}

%

\section{Introduction}
%
%
%
%
\IEEEPARstart{T}{eams} of autonomous robots that work cooperatively are used more and more widely for a range of robotic tasks \cite{MagEg01,BuCoMa09}. Robots have been deployed in formations with different shapes in order to facilitate the adaptive sampling of an unknown environment \cite{LePaLeSeFrDa07} or to achieve better cooperation efficiency \cite{MaStGrInSp08}.  As a result, considerable research efforts have been made in the past few years on designing distributed control laws to stabilize the shapes of formations of autonomous agents \cite{SuYa99,JaFre02,JoRa03,KwCh12}. In particular, within the research area of developing cooperative control theory for multi-agent systems, a sequence of theoretical investigations have been made to design formation control laws using the notion of graph rigidity \cite{AnYuFiHe08,KrBrFr08,YuAnDaFi09,CaYuAn11}, and such control laws are usually based on the gradients of the potential functions closely related to the graphs describing the distance constraints between the neighboring agents.

However, it has been recently reported in \cite{BeMoMoAn12, SunMouAndMor13} that for such gradient control laws, if agents disagree with their neighboring peers on the measured or prescribed distances between them, undesirable  formation motion might appear. Surprisingly, such  inconsistency induced motions take peculiar forms:   in $\R^2$, the agents move collectively in a distorted but rigid formation following a closed orbit that is determined by a single sinusoidal signal; in $\R^3$, the orbit becomes helical that is determined by a single sinusoidal signal and a constant drift. This is rather unexpected especially when knowing the robustness as a consequence of the \emph{exponential} convergence of gradient control; after all,  exponential convergence of a dynamical system usually implies its robustness against small disturbances. With the hindsight gained from \cite{BeMoMoAn12, SunMouAndMor13}, one realizes that the exponential convergence takes place for the error signals determined by the differences of the real and prescribed distances between neighboring agents, but this does not prevent the ill behavior of the position or velocity signals of the agents when measurement inconsistency exists. Such an observation is by no means trivial, but may affect the application of robotic formations because  robustness issue is particularly relevant in practice, where  distance disagreements may arise for several reasons. Firstly, robots may have different guidance systems, which may differ in their setting points; secondly, sensors equipped on robots may not return the same reading even if they are measuring the same distance due to heterogeneity in manufacturing processes; and thirdly, the same sensor can produce different readings for the same  distance in face of random measurement noises.

	In this paper, we focus on dealing with this tricky robustness issue by proposing to use an estimator-based gradient control. We are able to show that under mild assumptions, the proposed control strategy stabilizes formations in the presence of measurement inconsistency eliminating all the reported undesirable steady-state collective motions and distortion in the formations' final shapes. It takes full advantage of the strength of the existing gradient control, especially the exponential convergence speed, and at the same time preserves the distributed nature of the local cooperative control laws. We have discussed similar ideas in \cite{MaMiBa13} to install simple local estimators at the chosen estimating agents. We study more advanced estimator-based control in this paper that avoids the possible  high gains in control and handles a much broader class of measurement inconsistency. This inconsistency is in the form of a combination of a constant bias and a finite number of sinusoidal noise, which arises often in marine robotic tasks when sea waves perturb sensing	\cite{CaGa05} \cite{BeBaHo03}.


	The rest of the paper is organized as follows. In Section
	 II we describe the formation control problem for rigid formations in $\R^2$ and $\R^3$ and the robustness issues associated with gradient formation control. The estimator-based gradient control is proposed in Section III following the well established internal model principle rooted in robust control. In Section IV we carry out stability analysis and discuss in detail how to choose estimating agents systematically in Section V. Finally, in Section VI  experimental results are demonstrated using wheeled mobile robots moving in the plane and simulation results are discussed for mobile agents maneuvering  in the three dimensional space.

	\section{Rigid formations}
	We consider a formation in $\R ^m$, $m=2$ or $3$, consisting of $n\geq 2$ autonomous agents labeled by $1, \ldots, n$,  whose neighbor relationships are described by an undirected graph $\mathbb{G}$ with the vertex set $\mathcal{V}=\{1, \dots, n\}$ and the edge set $\mathcal{E} \subseteq \mathcal{V}\times \mathcal{V}$. We use $|\mathcal{E}|$ to denote the number of edges of $\mathbb{G}$. Let $k_{ij}$ denote the ordered pair $(i,j)$ to label the edge between vertices $i$ and $j$, and thus $k_{ij}\neq k_{ji}$. Let $\mathcal{E}_i$ denote the set of the labels in the form of $k_{ij}$ of all the edges associated with vertex $i$. To keep a desired shape of the formation, each agent $i$ is assigned with the task of keeping some prescribed distance $d_{k_{ij}}$ to every neighbor $j$. We assume that such distance constraints $d_{k_{ij}}$ are realizable in $\R^m$.

	Corresponding to the formation, $\mathbb{G}$ is embedded in $\R^m$ by assigning to each vertex $i$ a Cartesian coordinate $x_i \in \R^m$.
	A \emph{framework} is a pair $(\mathbb{G}, x)$, where $x = (x_1^T, \dots, x_n^T)^T$ is a multi-point in $\R^{mn}$. For every framework $(\mathbb{G}, x)$, we define the \emph{edge function} $f_{\mathbb{G}}
	: \R^{mn} \to \R^{|\mathcal{E}|}$ by
	\begin{equation}
		f_{\mathbb{G}}(x) = \operatorname*{col}_{\forall k_{ij}\in \mathcal{E}}(||z_{k_{ij}}||^2), \nonumber
	\end{equation}
	where $\operatorname{col}(\cdot)$ defines the column vector by collecting all its arguments as the vector's components, $z_{k_{ij}} = x_i - x_j$ is the relative position vector between vertices $i$ and $j$ for the edge $k_{ij}$ in the framework and $||\cdot||$ denotes the Euclidean norm.

	In order to define rigidity formations, we first review some basic notions on rigidity.
	\begin{definition}\cite{AsRo79} A framework $(\mathbb{G}, x)$ is \emph{locally rigid}
		if for every $x\in\R^{mn}$ there exists a neighborhood $\mathcal{X}$ of $x$ such that
	$f_{\mathbb{G}}^{-1}(f_{\mathbb{G}}(x)) \cap \mathcal{X}= f_{\mathbb{K}}^{-1}(f_{\mathbb{K}}(x))\cap \mathcal{X}$, where $\mathbb{K}$ is the complete graph with
	the same vertex set $\mathcal{V}$ of $\mathbb{G}$.\end{definition}
	\begin{definition}\cite{AsRo79} A framework $(\mathbb{G}, x)$ is \emph{globally rigid}
	if $f_{\mathbb{G}}^{-1}(f_{\mathbb{G}}(x)) = f_{\mathbb{K}}^{-1}(f_{\mathbb{K}}(x)).$\end{definition}

	Roughly speaking a framework $(\mathbb{G}, x)$ is rigid if it is not possible to smoothly move some
	vertices of the framework without moving the rest  while maintaining the edge lengths specified by
	$f_{\mathbb{G}}(x)$. If this property holds only locally in the neighborhood $\mathcal X$ of $x$, then the framework is only locally rigid; otherwise, if the property holds for the whole space, then the framework is globally rigid. Most of the existing literature has focused on a special class of rigid frameworks. We need some more definitions to introduce such frameworks.

	Let us take the following approximation of $f_{\mathbb{G}}(x)$
	\begin{equation}
		f_{\mathbb{G}}(x+\delta x) = f_{\mathbb{G}}(x) + df_{\mathbb{G}}(x) \delta x + O(\delta x^2), \nonumber
	\end{equation}
	where $df_{\mathbb{G}}(x)$ denotes the Jacobian matrix of $f_{\mathbb{G}}(x)$ and $\delta x$ is an infinitesimal displacement of $x$. The matrix $df_{\mathbb{G}}(x)$ is then called the \emph{rigidity matrix} of the framework $(\mathbb{G}, x)$.

	\begin{definition}\cite{AsRo79} A framework $(\mathbb{G}, x)$ is infinitesimally rigid
		 if $\operatorname{rank}df_{\mathbb{G}}(x) = 2n - 3$ in $\R^2$  or $\operatorname{rank}df_{\mathbb{G}}(x) = 3n - 6$ in $\R^3$ .
	\end{definition}

	Roughly speaking, an infinitesimally rigid framework $(\mathbb{G}, x)$ only admits rotations and translations of the whole framework in order
	to satisfy $f_{\mathbb{G}}(x+\delta x) = f_{\mathbb{G}}(x)$. The edge function remains constant up to the first order when $\delta x$ belongs to the kernel of $df_{\mathbb{G}}(x)$.

	Note that an infinitesimally rigid framework is also rigid, but in general the converse
	is not true. In order to state whether both frameworks are equivalent, we need
	to introduce the concept of regular points.

	\begin{definition}\cite{AsRo79} A multi-point $x$ is a \emph{regular point} of $(\mathbb{G}, x)$ if
	\begin{equation}
		\operatorname{rank}df_{\mathcal{G}}(x) = \operatorname{max}
		\{\operatorname{rank}df_{\mathbb{G}}(x) \, | \, x \in \R^{mn} \}. \nonumber
	\end{equation}
	\end{definition}

	\begin{theorem}\cite{AsRo79} A framework $(\mathbb{G}, x)$ is infinitesimally rigid if and only if
	$(\mathbb{G}, x)$ is rigid and $x$ is a regular point.
	\end{theorem}

	For an infinitesimally rigid framework $(\mathbb{G}, x)$ that is embedded in $\R^2$, it has at least $2n-3$ edges. If it has exactly $2n-3$ edges, then the framework is called \emph{minimally rigid}. For an infinitesimally rigid framework $(\mathbb{G}, x)$ that is embedded in $\R^3$, if it has exactly $3n-6$ edges then the framework is also called \emph{minimally rigid}.

	It is shown by Henneberg \cite{AnYuFiHe08}  that the 2D minimally rigid graphs on two or more vertices are exactly the graphs that can be obtained, starting from a single edge, by a sequence of operations of the following two types:
	\begin{enumerate}
		\item Add a new vertex to the graph, together with edges connecting it to two previously existing vertices.
		\item Subdivide an edge of the graph, and add an edge connecting the newly formed vertex to a third previously existing vertex.
	\end{enumerate}
	The first operation is referred to as the Henneberg insertion operation.

	In the next section, we discuss gradient control for rigid formations.

	\section{Gradient control and its robustness issue}
	Assume that agent $i$'s motion is described by a first-order kinematic  point model
	\begin{equation}\label{i agent}
	\dot x_i = u_i, \qquad i=1,\ldots, n,
	\end{equation}
	where $u_i \in \R^m$ is the control input for the agent $i$.

	In \cite{KrBrFr08}, an elegant distributed control law has been presented utilizing $z_{k_{ij}}$
	\begin{equation}
	u_i = -\sum_{k_{ij} \in \mathcal{E}_{i}}z_{k_{ij}}e_{k_{ij}},
	\label{eq: krick}
	\end{equation}
	where $e_{k_{ij}}$ is the error between the square of the real distance $\bar d_{k_{ij}}$ and the square of the prescribed distances $d_{k_{ij}}$ between the two agents $i$ and $j$ associated with edge $k_{ij}$
	\begin{equation}
		e_{k_{ij}} = ||\bar d_{k_{ij}}||^2 - d_{k_{ij}}^2.
	\label{eq: e def}
	\end{equation}
	It has been shown in \cite{KrBrFr08} that when $\bar d_{k_{ij}} = \bar d_{k_{ji}}$ and $d_{k_{ij}} = d_{k_{ji}}$, control law (\ref{eq: krick}) causes the solution of the closed-loop $n$-agent system to follow the direction of the gradient of the system's potential function $\frac{1}{4}\sum_{k_{ij} \in \mathcal{E}} e_{k_{ij}}^2$. Consequently, it is convenient to show that the errors $e_{k_{ij}}$ converge exponentially to zero when the formation is minimally rigid. For this reason, a number of  research groups have applied this gradient-based control law to a range of formation control problems under different settings \cite{YuAnDaFi09,CaYuAn11,CaMoYuAnDa11,OhAn11}.

	However, more recently, intriguing robustness issues of the gradient formation control have been reported in \cite{BeMoMoAn12,SunMouAndMor13}. For two neighboring agents $i$ and $j$, if there is some inconsistency in their measured or prescribed distance between them, namely $\bar d_{k_{ij}} \neq \bar d_{k_{ji}}$ or $d_{k_{ij}} \neq d_{k_{ji}}$,  and thus $e_{k_{ij}} \neq e_{k_{ji}}$, the control law (\ref{eq: krick}) does not correspond to the gradient of the potential function constructed in \cite{KrBrFr08} anymore. Indeed,  \emph{constant} inconsistency leads to two highly undesirable behaviors of the formation \cite{BeMoMoAn12,SunMouAndMor13}:
	\begin{enumerate}
	\item {\it Unknown distorted final shape.} When the inconsistency is small, the errors $e_{k_{ij}}$ converge to some unknown small but non-zero values, and thus the shape of the formation becomes distorted even as $t$ goes to infinity.
	\item {\it Steady-state collective motion induced by  inconsistency.} In $\R^2$, the agents move collectively in formation following a closed orbit that is determined by a single sinusoidal signal; in $\R^3$, the orbit becomes helical that is determined by a single sinusoidal signal and a constant drift.
	\end{enumerate}

	 Since measurement errors are ubiquitous in real robotic applications, this robustness issue inherent to the structure of gradient control poses urgent demand on designing new robust control strategies which preserves the exponential convergence property of gradient control and at the same time is robust against measurement discrepancies. One can show that the effects of $\bar d_{k_{ij}} \neq \bar d_{k_{ji}}$ and $d_{k_{ij}} \neq d_{k_{ji}}$ are equivalent in causing the undesired behavior just described. In this paper, to emphasize the possible measurement errors, we focus on deriving our system models for the case when $\bar d_{k_{ij}} \neq \bar d_{k_{ji}}$ while similar analysis carries over to the case when $d_{k_{ij}} \neq d_{k_{ji}}$.

	\section{Estimator-based gradient control}

	In this section, we present in detail how local estimators can be designed for chosen agents, called \emph{estimating agents}, such that measurement inconsistencies can be compensated distributively. Three main challenges are worth pointing out. First, the estimators' dynamics should not, if possible, affect the exponential convergence that is associated with the gradient control. Second, compensation should be done locally and different estimating agents should not give rise to conflicting compensation goals. Third, the class of discrepancies should be broad enough to contain at least the constant signals discussed in \cite{BeMoMoAn12,SunMouAndMor13}. In view of these challenges, one soon realizes that the design task is not easy at all. We have made some preliminary effort along this line in \cite{MaMiBa13}, where the estimator deals with only constant inconsistencies, and may run into high control gains. In what follows, we propose a novel estimator-based gradient control based on the well-known internal model principle that has been used for solving tracking and disturbance rejection problems \cite{ClBa13,ByPrIs97}, and more recently for cooperative control of  multi-agents systems \cite{WiSeAl11}.

	\subsection{Estimating agents}

	As we have discussed in the previous section, when there are distance measurement discrepancies, we have $\bar d_{k_{ij}} \neq \bar d_{k_{ji}}$ and thus $e_{k_{ij}} \neq e_{k_{ji}}$. We introduce the new variables $\mu_k$ for each edge $k = 1, \ldots, |\mathcal{E}|$ such that
	\begin{equation} \label{eq: mu}
	e_{k_{ij}} = e_{k_{ji}} - \mu_k.
	\end{equation}
	Obviously, the definition of $\mu_k$ distinguishes the two associated agents $i$ and $j$ since the indices $i$ and $j$ are not exchangeable in (\ref{eq: mu}). We call agent $i$, whose label is the leading subscript for edge $k$ on the left-hand side of (\ref{eq: mu}), the \emph{estimating agent} for edge $k$ since we will design an estimator for agent $i$ to estimate $\mu_k$ later. Then for each edge $k$, there is only one estimating agent associated with it.  We will discuss in Section VI how one chooses the estimating agents systematically. For each agent $i$, we use $\mathcal K _i$ to denote the set of the labels of the edges for which agent $i$ is chosen to be the estimating agent, and then $\mathcal K_i \subset \mathcal E_i$.

	\subsection{Modeling measurement inconsistency}
	We assume that the discrepancies $\mu_k$ are in the form of the superposition of a constant signal and $p$ sinusoidal signals with known frequencies $\{\omega_1, \omega_2, \dots, \omega_p\}$, namely 
	\begin{equation}
	\mu_k(t) = \alpha_k + \sum_{i=1} ^p \beta_i \sin (\omega_i t + \phi_i),
	\label{eq: noise model}
	\end{equation}
	where $\alpha_k$, $\beta_i$ and $\phi_i$ are fixed but unknown offset, amplitude and phase respectively. This noise model is widely used for formation control when the robots are known to work in the environment with periodic background noises. For example, short-term sea waves can be described by a superposition of periodic waves whose frequencies can be accurately estimated \cite{BeBaHo03}, and thus the measurement noise for underwater marine vehicles using floating buoys \cite{CaGa05} can be treated as the superposition of a finite number of sinusoidal signals with known frequencies.

	\subsection{Estimator-based control}
	We first propose the estimator-based gradient control. To explain the reasoning of the construction of the specific form of the estimator, we have to wait until we build up the state-space model for the overall closed-loop system.

	We propose to use the following distributed, estimator-based, dynamic, gradient control
	\begin{equation}
	 \label{eq: krick with muhat}
	 u_i = -\sum_{k_{ij} \in \mathcal{E}_{i}}z_{k_{ij}}e_{k_{ij}} + \sum_{k_{ij}\in \mathcal{K}_i}z_{k_{ij}}\hat\mu_{k},
	 \end{equation}
	where the first term is the same as the gradient control in (\ref{eq: krick}) and the second term uses $\hat{\mu}_k\in \R$ which is agent $i$'s estimate of the discrepancy $\mu_k$. This estimator's dynamics are described by
	\begin{align}
		\dot\xi_k &= \Lambda \xi_k + \kappa B_k(e_{k_{ij}} + \mu_k - \hat{\mu}_k) \label{eq: krick with muhat2}\\
		\hat\mu_k &= B_k^T\xi_k \quad  \label{eq: krick with muhat3}
	\end{align}
	where $\xi_k\in \R^{2p+1}$ is the state of estimator $k$ and it can be initialized arbitrarily,
	\begin{equation}
	\Lambda = \begin{bmatrix}0 & 0 & 0 \\ 0 & 0 & -\Omega\\
	0 & \Omega & 0\end{bmatrix}, \quad B_k = \begin{bmatrix}b_1 \\ b_2\end{bmatrix}, \label{eq: labk}
	\end{equation}
		$\Omega = \operatorname{diag}\{\omega_1, \omega_2, \dots, \omega_p \}$, the constants $b_1 \in \R$ and $b_2\in\R^{2p}$ are such that the pair $(B_k, \Lambda)$ is observable, and $\kappa > 0$ is  the gain to be designed.

	Now the first-order agent dynamics (\ref{i agent}), the estimator-based gradient control (\ref{eq: krick with muhat}) and the estimator dynamics (\ref{eq: krick with muhat2}) and (\ref{eq: krick with muhat3}) define precisely the closed-loop dynamics of each agent. To analyze the collective motion of the $n$-agent system, we need to build the state-space model.

	\subsection{State-space model of the closed-loop system}
	Consider all the $k_{ij} \in \cup \mathcal K_i$. Since for each edge in $\mathcal E$, there is only one estimating agent, we know that there are exactly $|\mathcal E|$ such $k_{ij}$. We stack all the corresponding $z_{k_{ij}}$, $e_{k_{ij}}$, $\mu_k$ and $\hat \mu_k$ together into column vectors to obtain the relative position,  error, inconsistency and estimation vectors $z = \operatorname*{col} (z_{k_{ij}}^T) \in \R^{m|\mathcal E|}$, $e = \operatorname*{col} (e_{k_{ij}}) \in \R^{|\mathcal E|}$, $\mu = \operatorname*{col} (\mu_{k}) \in \R^{|\mathcal E|}$ and $\hat \mu = \operatorname*{col} (\hat \mu_{k}) \in \R^{|\mathcal E|}$. Define the system's state $x = \matt{x_1^T & \cdots & x_n^T}^T \in \R ^{mn}$. Then the $n$-agent system dynamics derived from (\ref{i agent}) and (\ref{eq: krick with muhat}) are
	 \begin{align} \label{eq: all}
		 \dot x = -R(z)^T{e} - S_1^T(z)(\mu - \hat\mu),
	 \end{align}
	where $R(z) \dfb df_{\mathbb{G}}(x)$ is the rigidity matrix of graph $\mathbb G$, $S_1 = Z^T J$, $Z = \textrm{diag} \{z_{k_{ij}}\}$, $J$ is obtained by replacing all the $-1$ in $H\otimes I_m$ by zero, being $H$ the transpose of the incidence matrix of $\mathbb G$ \cite{Di97}.

	 Now we are ready to present the state-space model for the closed-loop $n$-agent system derived from, (\ref{eq: krick with muhat2}), (\ref{eq: krick with muhat3}) and (\ref{eq: all}). Note that the error system can be easily computed from (\ref{eq: all}) as $\dot e = 2R(z)\dot x$ as discussed in \cite{MaMiBa13}. More precisely, the closed-loop system can be written in the following compact form
	 \begin{align}
		\mathbf{P}&: \left\{
		\begin{array}{l l}
			\dot{ e} &= -2R(z)R(z)^T e - 2R(z)S_1(z)^T (\mu-\hat\mu) \\
		y &=  e
			\end{array}
			\right. \label{eq: 1e} \\
		\mathbf{C}&: \left\{
		\begin{array}{l l}
			\dot{\xi} &= M \xi + \kappa B(y + \mu - \hat\mu) \\
		\hat\mu &= B^T\xi
		\end{array}
		\right. \label{eq: 2mu} \\
		\mathbf{W}&: \left\{
		\begin{array}{l l}
		\dot w &= M w \\
		\mu &= B^Tw
		\end{array}
		\right., \label{eq: 3w}
	\end{align}
	where $\xi = \operatorname{col}\{\xi_k\}$ is the state of $\mathbf{P}$, $M=\operatorname{diag}\{\Lambda, \dots, \Lambda\}$, $B=\operatorname{diag}\{B_k\}$, and $w=\operatorname{col}\{w_k\}$ is the state of the exosystem $\mathbf{W}$ whose output is the discrepancy signal given in (\ref{eq: noise model}). Despite the fact that the variable $z$ appearing in (\ref{eq: 1e}) is a function of $x$ (which is not part of the state equation), it is worth to mention that the terms $R(z)R(z)^T$ and $R(z)S_1(z)^T$ can be expressed solely as a function of $e$ as discussed in \cite{MouMorseBelSunAnd14}. Hence the state equations (\ref{eq: 1e})-(\ref{eq: 3w}) defines an autonomous system. The initial estimates of the offset, phase and amplitude of $\mu_k$ are encoded in the initial condition $w(0)$. Note that the estimating agents are measuring $e+\mu$ as a whole, while the unknown $\mu$ appears in (\ref{eq: 2mu}). The signal flow of the closed-loop system is shown in the block diagram in Figure \ref{fig: im}.

	\tikzstyle{block} = [draw, fill=white, rectangle, 
		minimum height=3em, minimum width=6em]
	\tikzstyle{sum} = [draw, fill=white, circle, node distance=2cm]
	\tikzstyle{input} = [coordinate]
	\tikzstyle{output} = [coordinate]
	\tikzstyle{pinstyle} = [pin edge={to-,thin,black}]

	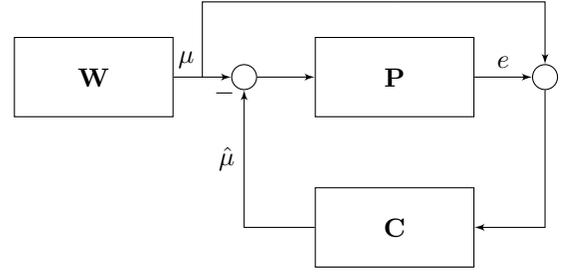
\begin{figure}
		\centering
	\begin{tikzpicture}[auto, node distance=2cm,>=latex']
		\node [block] (W) {$\mathbf{W}$};
		\node [sum, right of=W] (sum1) {};
		\node [block, right of=sum1] (P) {$\mathbf{P}$};
		\node [sum, right of=P] (sum2) {};
		\node [block, below of=P] (C) {$\mathbf{C}$};
		\node [coordinate, above of=P, node distance=1cm] (med) {};

		\path (W) -- coordinate (branch) (sum1);
		\draw [->] (W) -- node[name=mu, pos=0.22] {$\mu$} (sum1);
		\draw [->] (sum1) -- (P);
		\draw [->] (P) -- node {$e$}(sum2);
		\draw [->] (sum2) |- (C);
		\draw [->] (C) -| node[pos=0.99] {$-$} 
			node [near end] {$\hat\mu$} (sum1);
		\draw [-] (branch) |- (med);
		\draw [->] (med) -| (sum2);
	\end{tikzpicture}
	\caption{The plant $\mathbf{P}$ corresponds to the common error system $e$, where
	its input and output are perturbed by the discrepancy $\mu$ generated by the exosystem $\mathbf{W}$ as in (\ref{eq: 3w}). The disturbance rejection is achieved by using the internal controller $\mathbf{C}$ that is a copy of the exosystem.}
	\label{fig: im}
	\end{figure}

	Using standard framework in robust output regulation problem, one can take the inconsistency  $\mu$ to be the disturbances that directly influence the input and the output signals of the plant $\mathbf{P}$, and the controller $\mathbf{C}$ must contain internal models that are copies of the exosystem $\mathbf{W}$. One can check that in this case the Byrnes-Isidori regulator equation \cite{ByIs91} is solvable with the trivial solution $\xi = w$ and $e = 0$.

	After setting up the mathematical descriptions of the estimator-based control and the corresponding state-space system model, we are ready to show in the next section that the $n$-agent system under measurement inconsistency is exponentially stabilized by our proposed control.

	\section{Stability analysis}
	In the previous section, we have designed a distributed controller using the key idea of compensating the discrepancy  locally using internal-model-based estimators. Now we present our main result showing the performances of the proposed controller.
	\begin{figure*}[htp]
	\normalsize
	\begin{equation}
		\label{eq: big}
		\begin{bmatrix}
		\dot{e} \\ \dot\alpha \\ \dot\theta
		\end{bmatrix} =
		\begin{bmatrix}
		-2R(z)S_2(z)^T & -2R(z)S_1(z)^T & 0 \\
			-2R(z)S_2(z)^T & -2R(z)S_1(z)^T -\kappa B^TB & B^TM \\
		-\kappa B & 0 & M - \kappa BB^T
		\end{bmatrix}
		\begin{bmatrix}
		e \\ \alpha \\ \theta
		\end{bmatrix}
	\end{equation}
	\hrulefill
	\end{figure*}
	\begin{theorem}
	For the closed-loop $n$-agent formation (\ref{eq: 1e}) and (\ref{eq: 2mu}) with the measurement inconsistency vector $\mu$ driven by the exosystem (\ref{eq: 3w}) and unknown initial condition $w(0)\in\R^{(1+2p)|\mathcal{E}|}$, if the matrix
	\begin{equation}
	Z \dfb -R(z^*)R(z^*)^T + R (z^*) S_1(z^*)^T
	\end{equation}
	is Hurwitz at the desired relative position $z^*$ corresponding to $e=0$, then there exist  $\kappa^* > 0$ and $b^*>0$ such that for any $\kappa ||B_k||^2 > \kappa^*$ and $||b_2|| < b^*$, the system admits a locally exponentially attractive invariant manifold $\mathcal Q \dfb \{ (w, x,\xi) \, | \, e = 0, \xi = w \}$, and thus the  shape of the formation  converges exponentially to the desired shape defined by $e = 0$, the estimation $\hat\mu$ converges exponentially to $\mu$ and the velocity $\dot x$ converges exponentially to zero (i.e. the formation eventually stops).

	\label{thm: main}
	\end{theorem}
	\begin{IEEEproof}
		We take the coordinate transformation $\alpha = e + \mu - B^T\xi$ and $\theta = w-\xi$, and then the  equations (\ref{eq: 1e})-(\ref{eq: 3w}) can be rewritten into (\ref{eq: big}), where $S_2(z) = R(z) - S_1(z)$. We then calculate its Jacobian matrix $L$ at the equilibrium point $e=0$, $\alpha = 0$ and $\theta =0$. Although the system matrix of (\ref{eq: big}) is state dependent, several of its components are functions of the inner products $z_i^T z_j$ and thus their partial derivatives can be computed straightforwardly. In fact, the Jacobian matrix is
	\begin{equation}
	L = \begin{bmatrix}L_1 & L_2 \\ L_3 & L_4 \end{bmatrix},
	\end{equation}
	where
	\begin{align}
		L_1 &= \begin{bmatrix}
	-2R(z^*)S_2(z^*)^T & -2R(z^*)S_1(z^*)^T \\
	-2R(z^*)S_2(z^*)^T & -2R(z^*)S_1(z^*)^T -\kappa B^TB
	\end{bmatrix} \nonumber \\
		L_2 &= \begin{bmatrix}
		0 \\ B^TM
	\end{bmatrix}, \quad
	L_3 = \begin{bmatrix}
		-\kappa B & 0
	\end{bmatrix}, \quad
	L_4 = M - \kappa BB^T. \nonumber
	\end{align}
	We now prove that $L_1$ is Hurwitz, and this is equivalent to prove the system
	\begin{align}
		\dot{e} &= -2R(z)S_2(z)^Te -2R(z)S_1(z)^T\alpha  \label{eq: n1} \\
		\dot\alpha &= -2R(z)S_2(z)^Te -2R(z)S_1(z)^T\alpha -\kappa B^TB\alpha
		\label{eq: n2}
	\end{align}
	is asymptotically stable at the origin $e=0$ and $\alpha = 0$ in which case $\hat \mu = \mu$. Let $ f(e, \alpha) \dfb -2R(z)S_2(z)^Te -2R(z)S_1(z)^T\alpha$, and we compute \begin{equation}
		F_0 \dfb {\frac{\partial f(e, 0)}{\partial e}} |_{e = 0} = 2Z,
	\label{eq: as1}
	\end{equation}
	which is Hurwitz since $Z$ is in view of the condition in the theorem.
	Therefore, there exists a positive definite matrix $P = P^T$ such that
	\begin{equation}
		F_0^TP+PF_0 = -2I.
		\label{eq: FP}
	\end{equation}
	Then for system (\ref{eq: n1})-(\ref{eq: n2}) consider the candidate Lyapunov function
	\begin{equation}
	V(e, \alpha) = e^TPe + \frac{1}{2}\alpha^T\alpha,
	\end{equation}
	whose time derivative along the system's solution is
	\begin{align}
		\dot V(e, \alpha) &= 2e^TPf(e, \alpha) + \alpha^T
	f(e, \alpha) - \kappa ||B||^2 ||\alpha||^2 \nonumber \\
	&= 2e^TPf(e, 0) +  2e^TP\left(f(e, \alpha)-f(e, 0)\right) + \nonumber \\
	 & \qquad + \alpha^T f(e, \alpha)- \kappa ||B||^2 ||\alpha||^2 \nonumber \\
	&= 2e^TP(F_0 e + g(e) + \bar f(e)\alpha)
	+ \alpha^T f(e, \alpha) \nonumber \\
	  & \qquad- \kappa ||B||^2 ||\alpha||^2,
	\label{eq: Vder}
	\end{align}
	where $g(e)$ is the Taylor-series residue that satisfies
	\begin{equation}
	\lim_{||e|| \to 0} \frac{||g(e)||}{||e||} = 0,
	\label{eq: lim}
	\end{equation}
	and $\bar f(e) = -2R(z)S_1(z)^T$. In particular, (\ref{eq: lim}) implies that for any $\epsilon > 0$, there exists a $\delta$ such that $||e|| \leq \delta \implies ||g(e)||\leq \epsilon ||e||$.
	Taking $\epsilon = \frac{1}{2||P||}$ with the corresponding $\delta$, since $\bar f(e)$ is locally Lipschitz, we know that for all $||e|| \leq \delta$, $||\alpha|| \leq \delta$, there exist $p, q > 0$ such that
	\begin{align}
		\dot V(e, \alpha) &\leq -2||e||^2 + 2||e||||P||\frac{1}{2||P||}||e|| + p||e||||\alpha|| + \nonumber \\
		& + q||\alpha||^2 -\kappa ||B_k||^2 ||\alpha||^2,
	\end{align}
	where the third and fourth terms on the right-hand side are due to the boundedness of $\bar f(e)$ in an open ball. Hence, by choosing
	$\kappa ||B||^2 > 0$ such that
	\begin{equation}
		q + \frac{p^2}{2} -\kappa ||B||^2 \leq -\frac{1}{2},
	\end{equation}
	we have, in view of Young's inequality, that
	\begin{equation}
	\dot V(e, \alpha) \leq -\frac{1}{2}(||e||^2 + ||\alpha||^2),
	\end{equation}
	and thus system (\ref{eq: n1})-(\ref{eq: n2}) converges exponentially to the origin for all the initial conditions $e(0), \alpha(0)$ starting in the set $\mathcal{C}:=\{x, \hat\mu, \mu \,|\,  \lambda_{\text{min}}(P)||e||^2 + \frac{1}{2}||e + \mu - \hat\mu||^2 \leq \delta^2 \}$. So we have proved that $L_1$ is Hurwitz.

	We further observe that $L_4$ is Hurwitz for any $\kappa > 0$, which follows from the  asymptotic stability of $\dot\theta = L_4\theta$ since $L_4+L_4^T = -2\kappa BB^T$ and the pair $(B, M)$ is observable. Thus, if $L_2$ is zero, i.e. $b_2 = 0$ since $B_k^T = \begin{bmatrix}b_1 & \bf{0}\end{bmatrix}$ is a left eigenvector for the single zero eigenvalue of $\Lambda$, then the eigenvalues of $L$ are the eigenvalues of $L_1$ and $L_4$. Therefore, for a sufficiently small $B$ such that $0 < ||b_2|| < b^*$, $L$ is still Hurwitz. Hence,  system ($\ref{eq: big}$) is locally exponentially stable, which implies that $(w, x, \xi)$ locally exponentially converges to $\mathcal{Q}$. Since $\dot x = 0$ in the invariant manifold $\mathcal{Q}$, we conclude also that $\dot x(t)\to 0$ exponentially as $t\to\infty$, i.e. the formation eventually stops.
	\end{IEEEproof}

	\begin{remark}
		For the sake of clarity, we have assumed that $M$ is the same for all the inconsistencies $\mu_k$. It can be checked that the result in Theorem \ref{thm: main} still holds for having different sets of frequencies for each inconsistency $\mu_k$. Note that we have not only removed all undesired effects induced by the presence of inconsistency, but with the estimation of $\mu(t)$, Theorem \ref{thm: main} provides a systematic method to calibrate the offset of the sensors in the estimating agents with respect to the sensors in the non-estimating agents.
	\end{remark}

	In the next section, we explain how to choose the estimating agents systematically to guarantee the conditions in   Theorem \ref{thm: main} to hold.

	\section{Selecting the estimating agents}
	The condition of $Z$ being Hurwitz in Theorem \ref{thm: main} is a sufficient condition for the local exponential stability of system (\ref{eq: 1e})-(\ref{eq: 2mu}). To check this condition, one needs to calculate the eigenvalues of an $|\mathcal{E}| \times |\mathcal{E}|$ square matrix. Such computations can be burdensome and in this section we are going to show that for a large class of infinitesimally minimally rigid formations one can still guarantee the admissibility of the condition by choosing smartly the estimating agents and thus avoid computing the eigenvalues.

	\subsection{Stabilizing a large class of infinitesimally minimally rigid  formations in $\R^2$}

	In this subsection we study a class of infinitesimally minimally rigid formations in $\R^2$ that are generated by a sequence of Henneberg insertion operations starting from triangular formations, for which we present two ways of picking the estimating agents. Then we introduce a systematic way of choosing the estimating agents based on the Henneberg insertion described at the end of Section II. We remark that a range of minimally rigid formations can be generated through the Henneberg insertion operation \cite{Haas05}.

\begin{proposition}
	\label{pro: triang cycle}
For any undirected triangular formation, where each agent acts as an estimating agent for only one edge, then its associated $Z$ matrix is Hurwitz.
\end{proposition}
\begin{figure}
	\centering
	\includegraphics[width=0.18\textwidth]{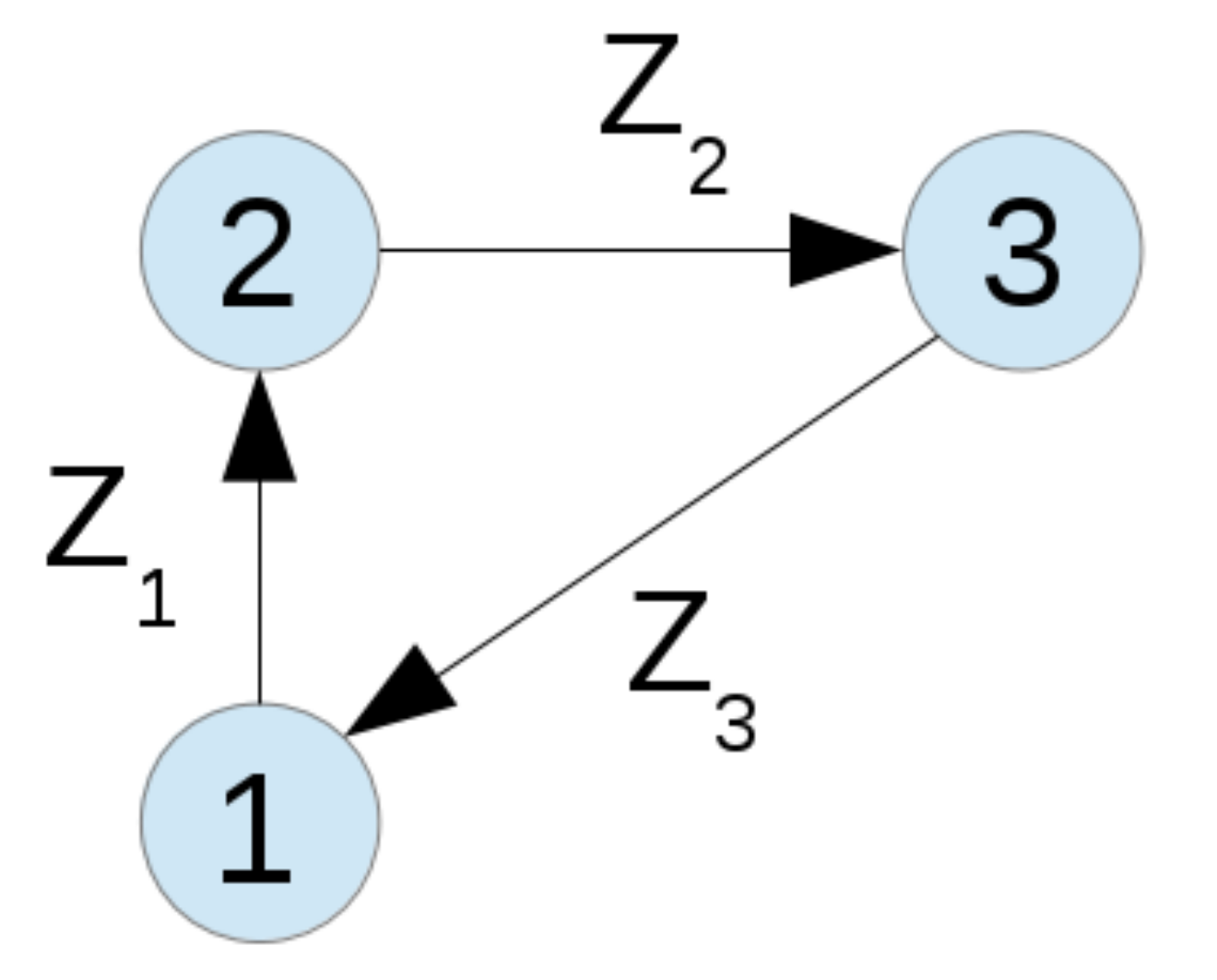}
	\caption{A triangular formation associated with a cyclic estimating-agent graph where the tails of the arrows indicate the corresponding estimating agents. }
	\label{fig: j triang}
\end{figure}

\begin{IEEEproof} One can check that in this case $R(z)S_2(z)^T + S_2(z)R(z)^T = R(z)R(z)^T$. In addition,  $R(z^*)R(z^*)^T$ is  positive definite matrix since undirected  triangular formations are minimally rigid. So $-R(z)S_2(z)^T$ is Hurwitz at $z=z^*$ or equivalently $e = 0$, and this in turn is equivalent to $Z$ is Hurwitz.
\end{IEEEproof}

\begin{proposition}
	\label{pr: 3acy}
For any undirected triangular formation, where one agent is the estimating agent for both of the two edges that it is associated with and exactly one other agent is the estimating agent for the remaining edge, then its $Z$ matrix is Hurwitz.
\end{proposition}

\begin{IEEEproof}
In this case, we have
\begin{equation}
	R(z)S_2^T(z) = \begin{bmatrix}
		||z_{1_{12}}||^2 & 0 & 0 \\
		-z_{2_{23}}^Tz_{1_{12}} & ||z_{2_{23}}||^2 & -z_{2_{23}}^Tz_{3_{31}} \\
		0 & -z_{3_{31}}^Tz_{2_{23}} & ||z_{3_{31}}||^2
	\end{bmatrix},
	\label{eq: rs2}
\end{equation}
which can be rewritten into the block lower-triangular form
\begin{equation}
	R(z)S_2^T(z) = \begin{bmatrix}
	A_{11} & 0 \\
		A_{12} & A_{22}
	\end{bmatrix}.
\end{equation}
Here, $A_{11} = ||z_{1_{12}}||^2$ is always positive; the characteristic polynomial of $A_{22}$ is quadratic and thus it is easy to compute that both of its two eigenvalues live in the open left half-plane when $z_{2_{23}}$ and $z_{3_{31}}$ are not parallel, which has to be true since the formation is rigid. So the $Z$ matrix itself is Hurwitz.
\end{IEEEproof}

The two situations of choosing estimating agents for triangular formations are illustrated in Fig. 2 and Fig. 3 respectively.

\begin{figure}
	\centering
	\includegraphics[width=0.18\textwidth]{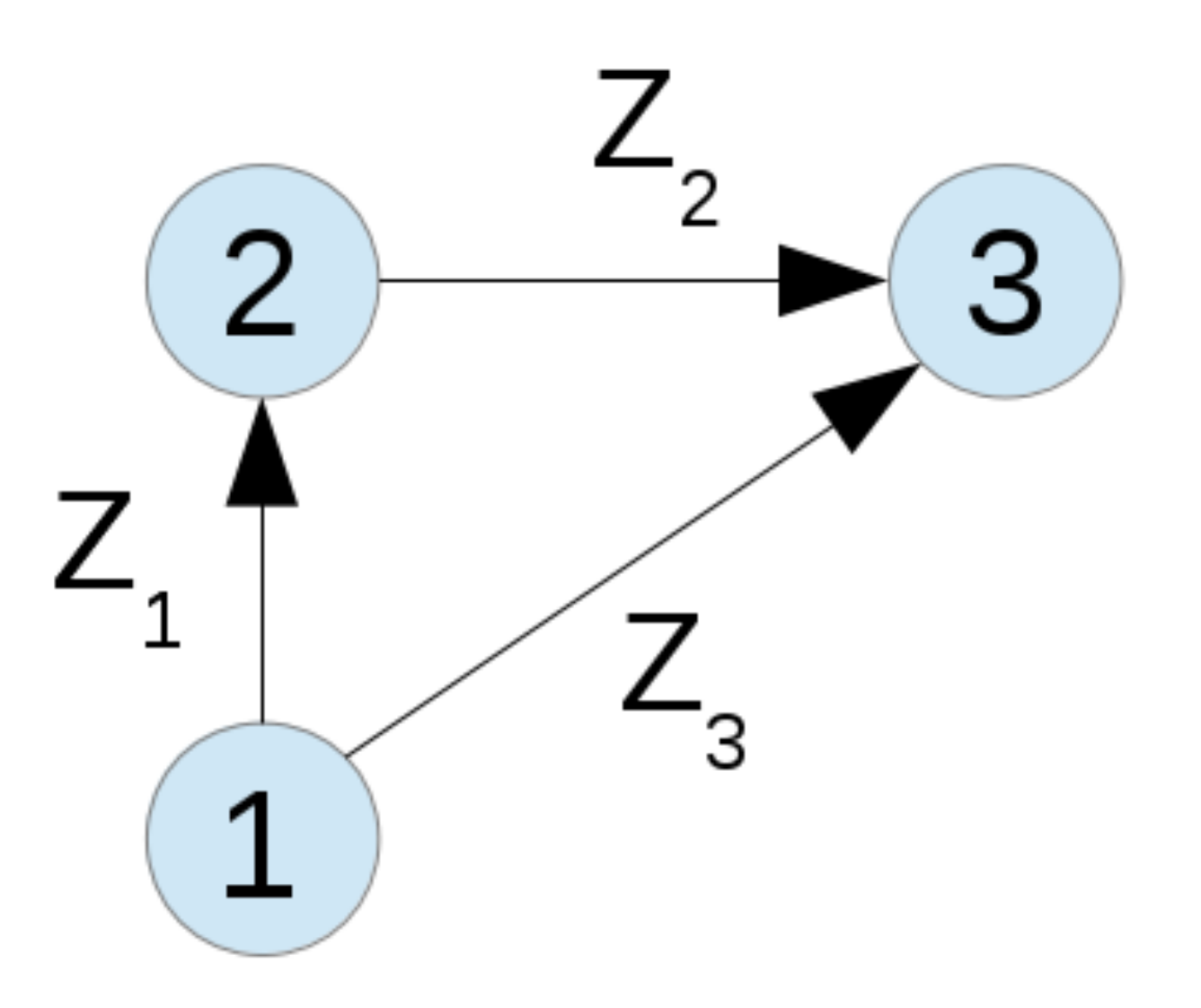}
	\caption{A triangular formation associated with an acyclic estimating-agent graph where the tails of the arrows indicate the corresponding estimating agents.}
	\label{fig: j triang2}
\end{figure}

\begin{proposition} 	\label{pro: nHen}
	For any infinitesimally minimally rigid formations in $\R^2$ that is generated by the Henneberg insertion operation, its associated $Z$ matrix is Hurwitz if one chooses the estimating agents following exactly the sequence of the Henneberg insertions and in addition: (i) for the first three agents, pick the estimating agents as in Proposition \ref{pro: triang cycle} or Proposition \ref{pr: 3acy}; (ii) for any new insertion operation that has just added two edges from a new agent to two existing agents, pick those two existing agents to be the estimating agents for the newly added two edges. Note that only those two chosen estimating agents are involved and the other agents are not affected at all.
\end{proposition}

\begin{IEEEproof} It suffices to prove that for an $n$-agent, $n\geq 3$, minimally rigid formation in $\R^2$ whose $Z$ matrix is Hurwitz with its chosen estimating agents, the new $(n+1)$-agent formation obtained from the $n$-agent formation by the Henneberg insertion operation still has a Hurwitz $Z$ matrix if its estimating agents are chosen according to the rule stipulated in the proposition.

Let the $n$-agent formation's $Z$ matrix be $R(z)S_2^T(z)$ and the $Z$ matrix for the $(n+1)$-agent formation be $\bar R(z)\bar S_2^T(z)$. Then
\begin{equation}
	\bar R(z) \bar S_2^T(z) =
\begin{bmatrix}
	R(z)S_2^T(z) & 0 \\
	\divideontimes & C(z)
\end{bmatrix},
\label{eq: rs2hh}
\end{equation}
where ``$\divideontimes$" denotes the submatrix of less importance and
\begin{equation}
C(z) = \begin{bmatrix}
	||z_l||^2 & - z_l^Tz_{l+1} \\
	-  z_{l+1}^Tz_l & ||z_{l+1}||^2
	\end{bmatrix},
\end{equation}
where $z_l$ and $z_{l+1}$ are the two vectors pointing from the two chosen estimating agents' positions to the $(n+1)th$ agent's position. Then using the similar argument as proving Proposition \ref{pr: 3acy}, one can show that $\bar R(z) \bar S_2^T(z)$ is also Hurwitz.
\end{IEEEproof}

\subsection{Stabilizing a special class of infinitesimally minimally rigid  formations in $\R^3$}
Now we look at undirected rigid formations in $\R^3$. We start with simple tetrahedron formations.
\begin{figure}
	\centering
	\includegraphics[width=0.18\textwidth]{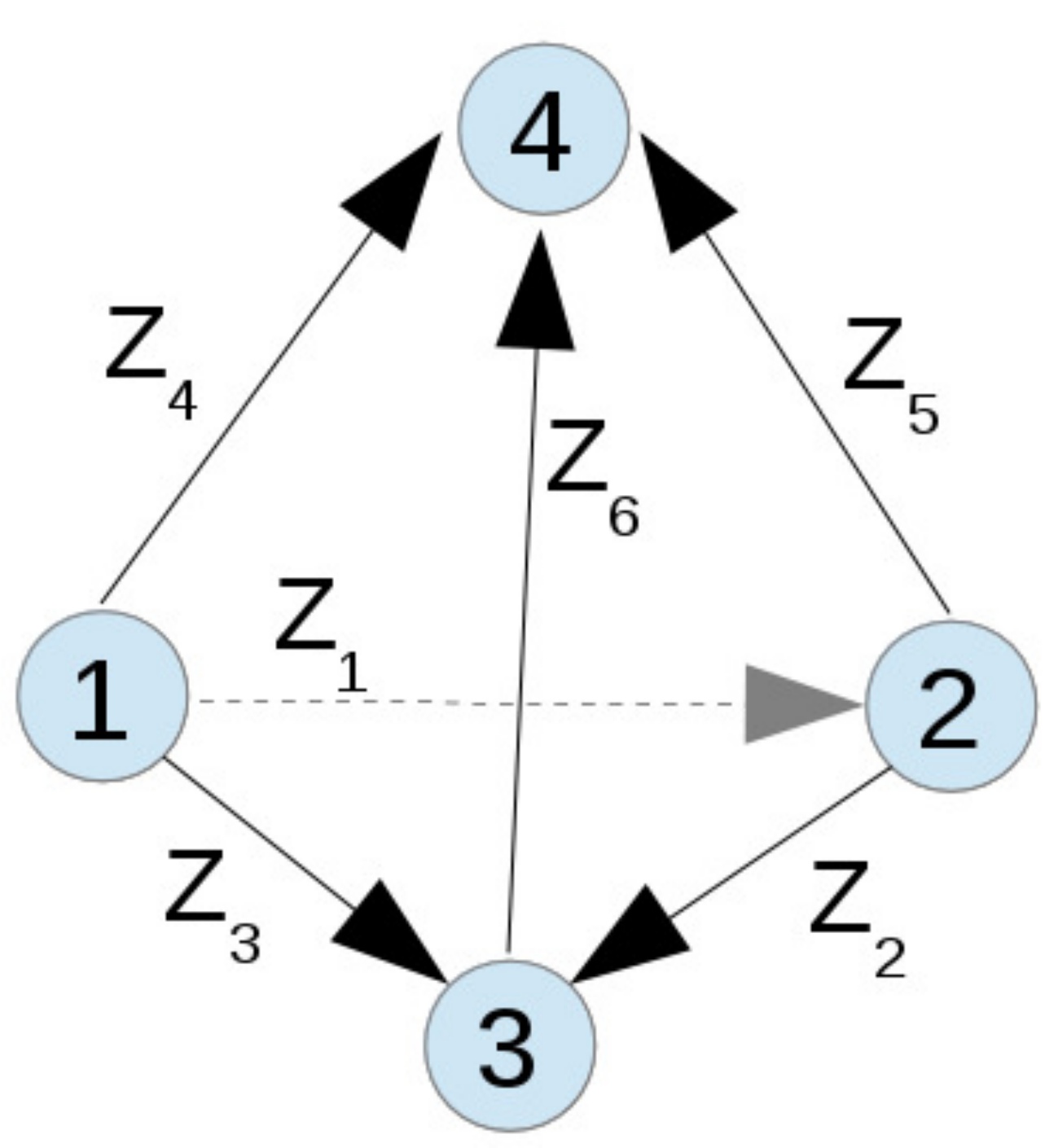}
	\caption{Tetrahedron formation where the tails of the arrows	indicate the corresponding estimating agents.}
	\label{fig: tetra}
\end{figure}
\begin{proposition}
	\label{pro: 3d}
For an undirected tetrahedron formation in $\R^3$, if its estimating agents are chosen as shown in Figure \ref{fig: tetra}, then its $Z$ matrix is Hurwitz.
\end{proposition}
\begin{IEEEproof}
The $Z$ matrix for the tetrahedron formation with the estimating agents chosen as shown in Figure \ref{fig: tetra} is
\begin{equation}
R(z)S_2^T(z) = \begin{bmatrix}A(z) & 0 \\ \divideontimes & G(z)\end{bmatrix},
	\label{eq: rs2 tetra}
\end{equation}
where $A(z)$ is the same matrix as in (\ref{eq: rs2}) and $G(z)$ is the Grammian matrix $\bar z^T\bar z$ and $\bar z$ is the stacked column vector of $z_4$, $z_5$, and $z_6$. Since the tetrahedron formation is minimally rigid at $z^*$, all the vectors in $\bar z$ are linearly independent. Therefore, $G(\bar z)$ is positive definite at $z^*$ and thus the $Z$ matrix is Hurwitz.
\end{IEEEproof}

Since there is no necessary and sufficient combinatorial conditions for formations' rigidity properties in $\R^3$ yet, we can only look at a special class of rigid formations in $\R^3$.
\begin{proposition}
	\label{pro: 3dn}
Consider the class of infinitesimally minimally rigid formation in $\R^3$ that can be generated by adding in sequence new agents to a tetrahedron formation such that every time the new agent is connected to three existing agents that form a triangular formation. If in each insertion operation, the three estimating agents for the three newly added edges are exactly the three associated existing agents, then the $Z$ matrix for the overall formation is Hurwitz.
\end{proposition}
The proof for this proposition is similar to that of Proposition \ref{pro: nHen} and we omit it here.

\section{Experimental and simulation results}
\subsection{Formation experiments in $\R^2$}
\begin{figure}
\centering
\includegraphics[width=0.3395\textwidth]{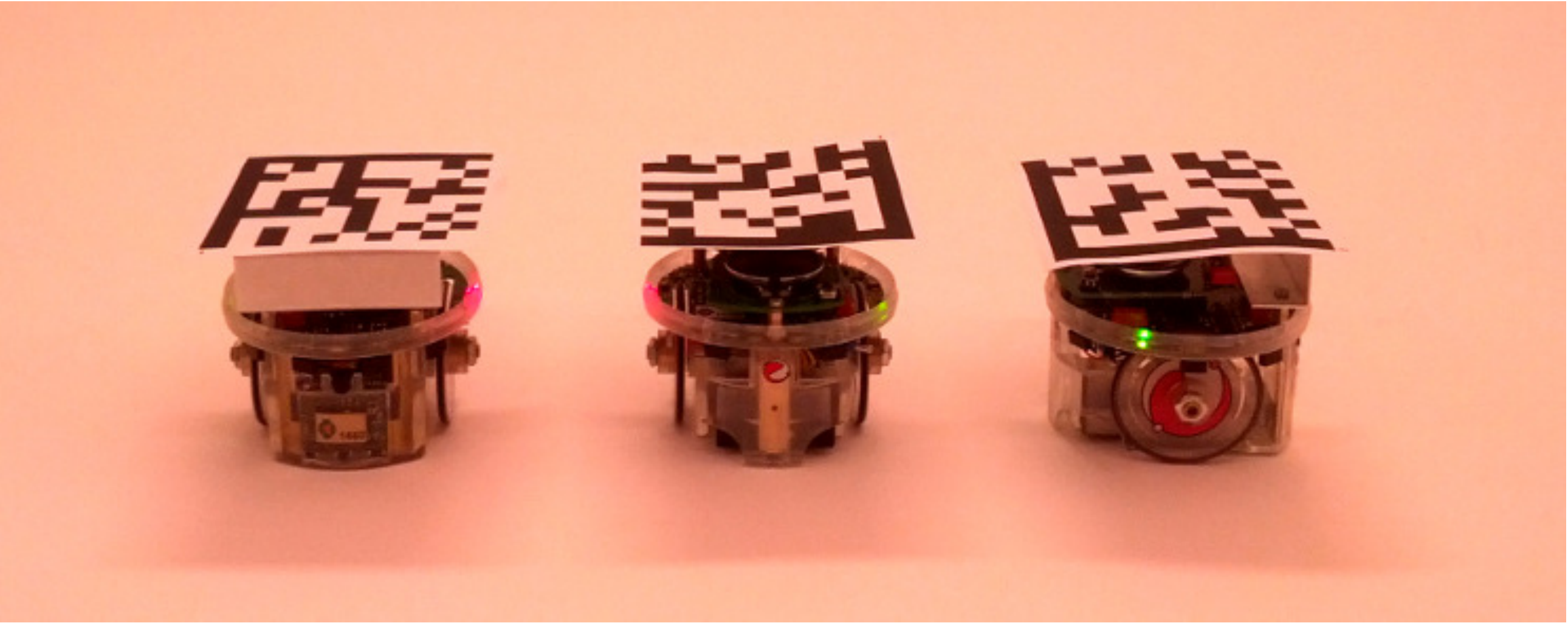}
\caption{Three wheeled E-pucks with data-matrices as markers on their tops that are used in the experimental setup.}
\label{fig: epucks}
\end{figure}

We first test the result in Theorem \ref{thm: main}
using the E-puck mobile robotic platform \cite{epucks}. The experimental setup consists of four
wheeled E-puck robots in a planar area of $2.6 \times 2$ meters.
Each robot is identified by a data-matrix marker on
its top as shown in Figure \ref{fig: epucks}. Each robot's reference point is the intersection of  the two solid bars of the marker  and the orientation of the marker is recognized by a vision algorithm running at a PC connected to an overhead camera. Since E-pucks are usually modeled by unicycles,
we apply feedback linearization about their reference points
to obtain single-integrator dynamics for simpler controller
implementation. In essence, we control the formation of the  reference points of the robots. The whole image
of the testing area is covered by $1600 \times 1200$ pixels, where
the distance between two consecutive horizontal or vertical
pixels corresponds approximately to $1.6$mm. The PC runs
a real time process computing the relative vectors between
the robots and computes the
control inputs for the robots. The
communication takes place when sending the commands
from the PC to the E-pucks in order to move their wheels, which gives the required linear and angular
velocities to the robots after being translated into
common (linear velocity) and differential (angular velocity)
commands to the wheels of the robots. The communication
is done via Bluetooth at the fixed frequency of 20Hz.

We consider the following fixed measurement inconsistency that is unknown to the robots
\begin{equation}
\mu = \begin{bmatrix}19 & 16 & 19.5 & 10 & 16 \end{bmatrix}^T \text{pixels}^2.
	\label{eq: muexp}
\end{equation}
The magnitude of such inconsistency is carefully chosen to reflect the possible bias of about $\sqrt{16}\times 1.6 = 6.4$mm, which is quite common among acoustic or infrared sensors for this kind of robots.

\begin{figure}
\centering
\begin{subfigure}{0.128\textwidth}
\includegraphics[width=\textwidth]{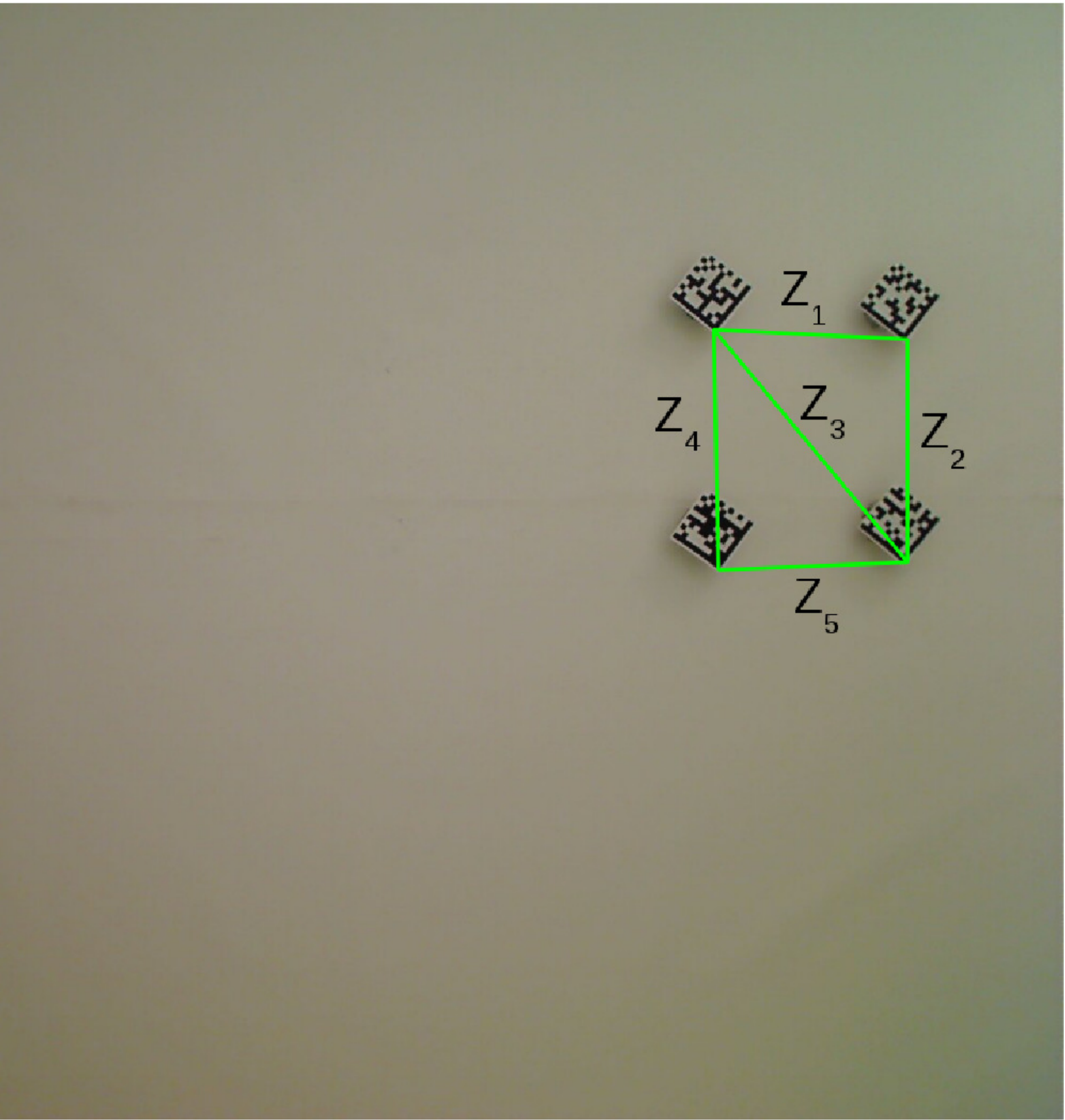}
\caption{}
\end{subfigure}
\begin{subfigure}{0.128\textwidth}
\includegraphics[width=\textwidth]{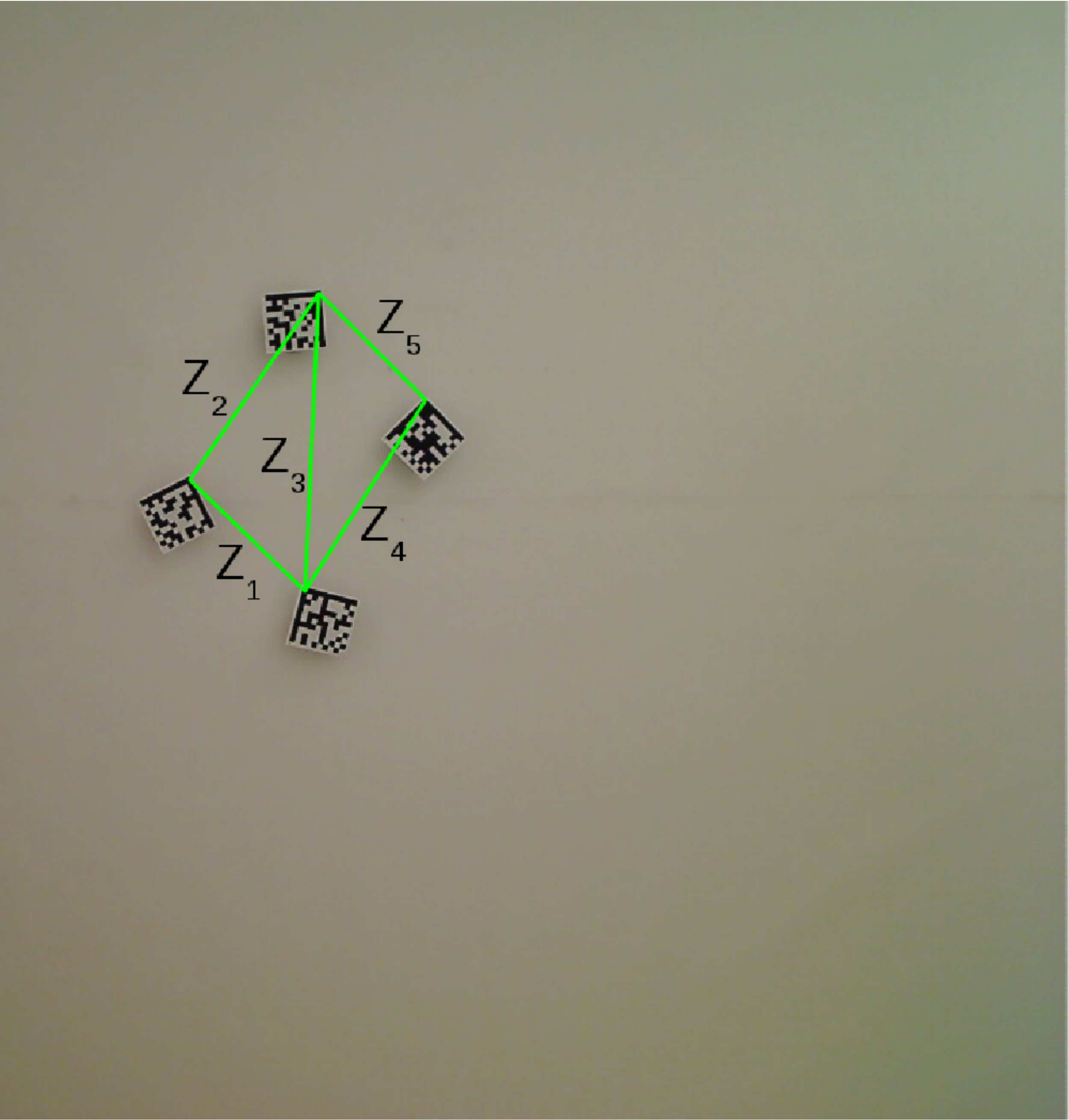}
\caption{}
\end{subfigure}
\begin{subfigure}{0.218\textwidth}
\includegraphics[width=\textwidth]{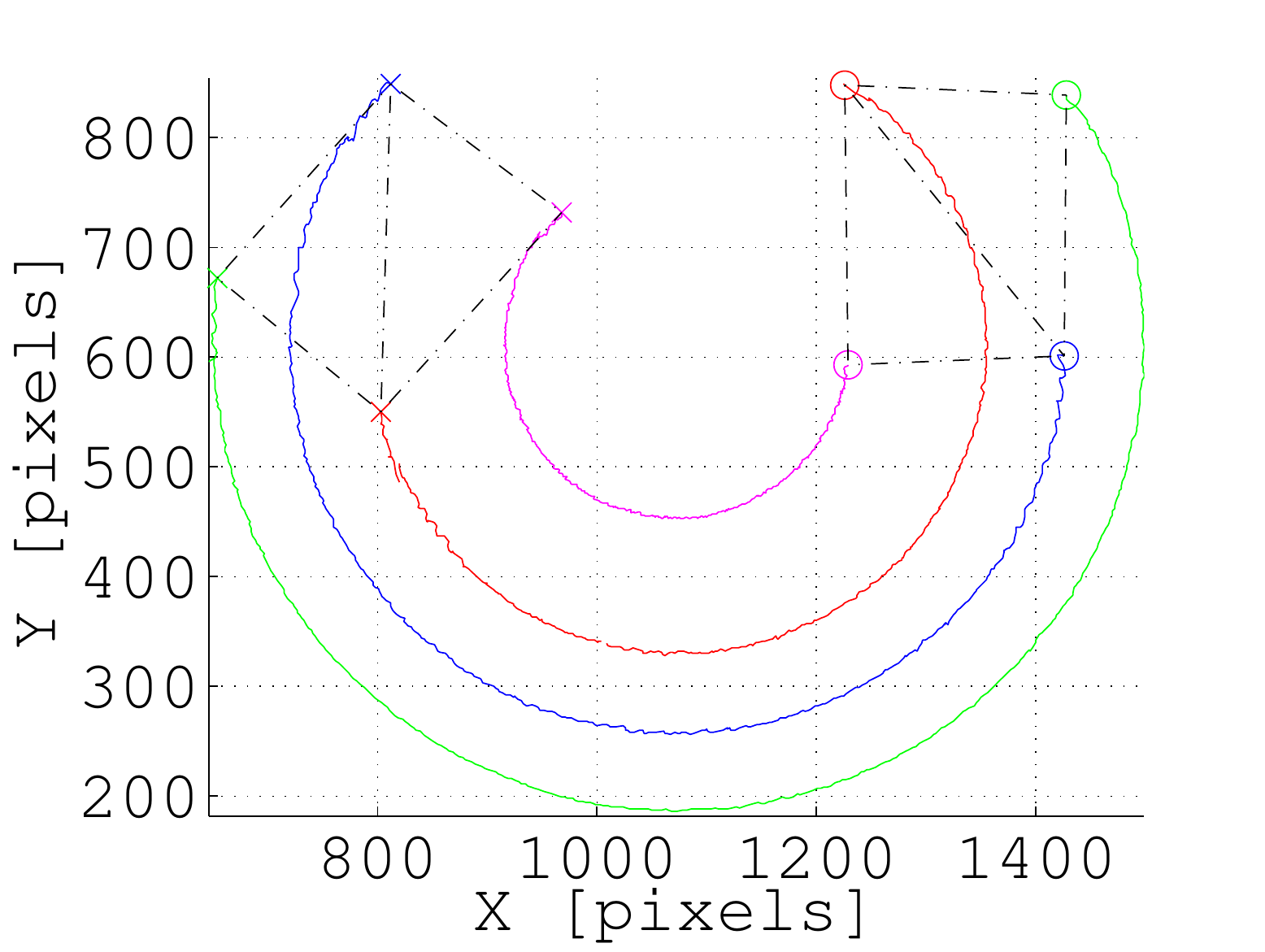}
\caption{}
\end{subfigure}
\begin{subfigure}{0.241\textwidth}
\includegraphics[width=\textwidth]{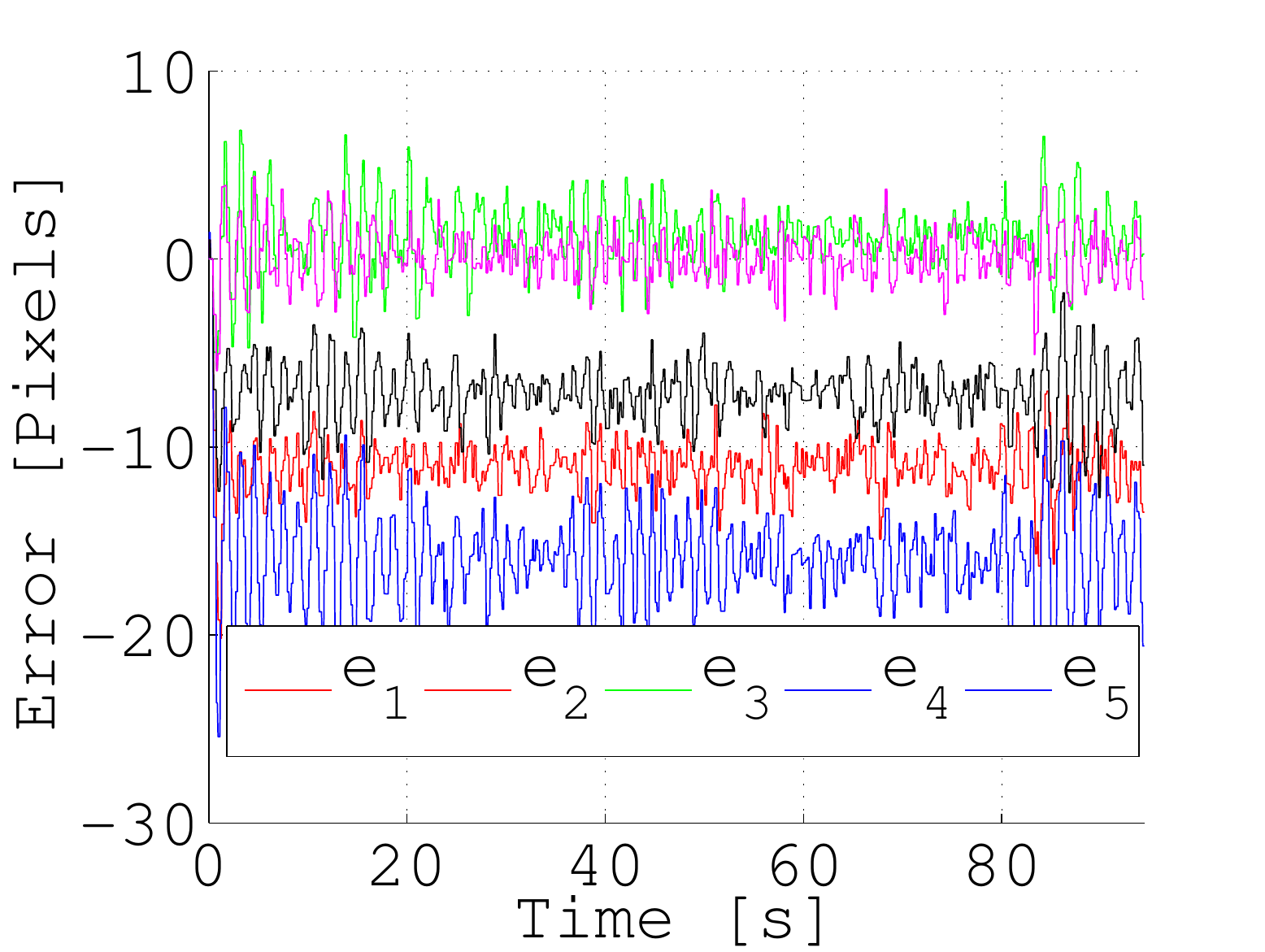}
\caption{}
\end{subfigure}
\begin{subfigure}{0.241\textwidth}
\includegraphics[width=\textwidth]{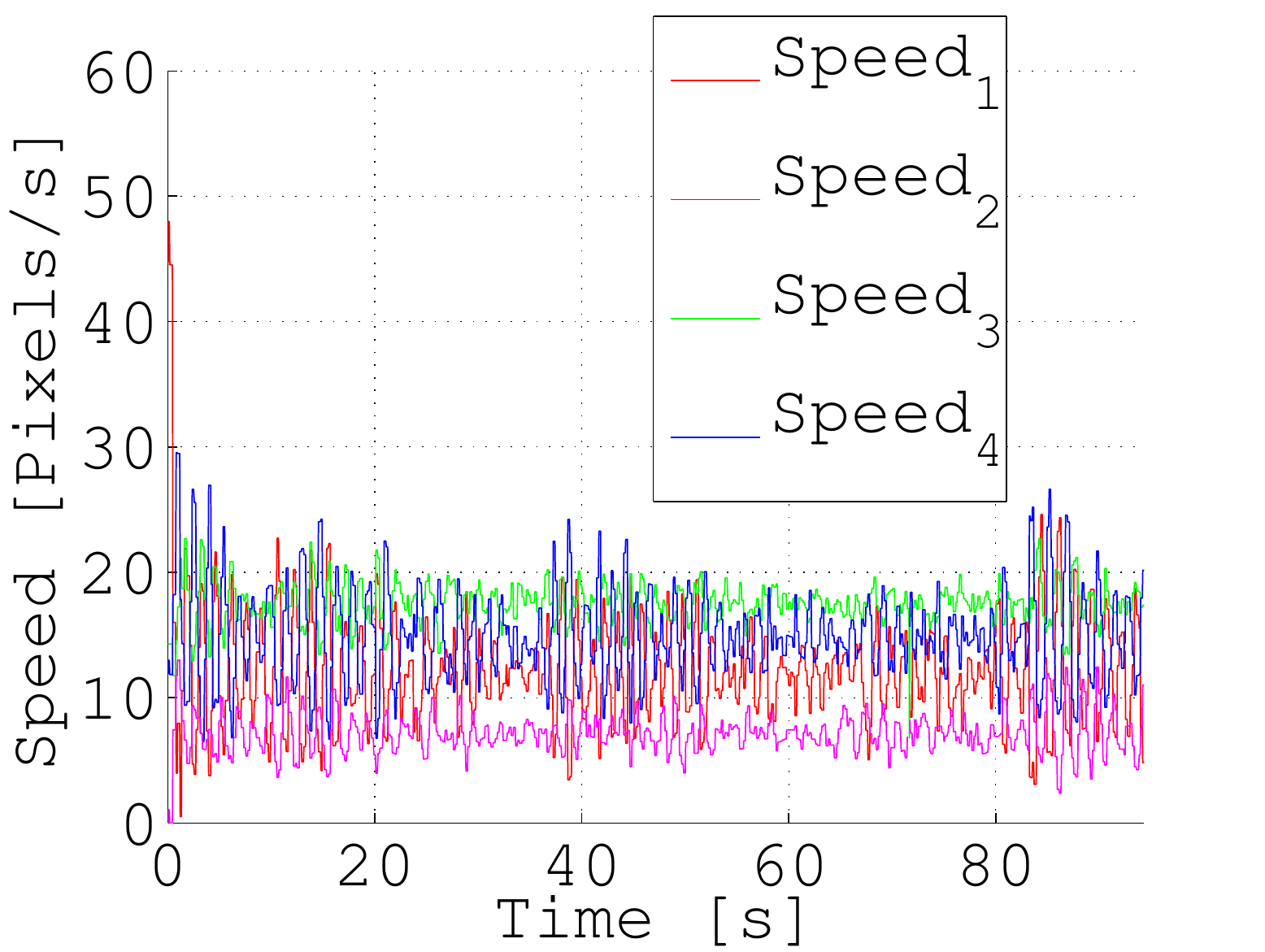}
\caption{}
\end{subfigure}
\caption{Experimental results of a 2D formation with inconsistent measurements where a standard gradient-based control is used (i.e. without the use of our proposed estimator); (a) initial configuration; (b) final configuration after 100 seconds; (c) the robot's trajectories; (d) the plot of errors $e_k$, where $k = 1,\dots, 5$; (e) speeds of the four robots which show that the robots never stop.}
\label{fig: dis}
\end{figure}

When the robots use directly the standard gradient control strategy, the measurement inconsistency induces the closed orbit as shown in Figure \ref{fig: dis} and the shape of the formation is distorted.  In comparison, for the same setup, we also apply the estimator-based gradient control  (\ref{eq: krick with muhat})-(\ref{eq: krick with muhat3}). We pick the estimating agents following the rule specified by Proposition 5.3 and as a result the transpose of the incidence matrix of the associated estimating-agent graph is
\begin{equation}
	H = \begin{bmatrix}
		1  &  -1  &   0  &   0\\
		0  &   1  &  -1  &   0\\
		1  &   0  &  -1  &   0\\
		-1  &   0  &   0  &   1\\
		0  &   0  &   1  &  -1
	\end{bmatrix}.
\end{equation}
We choose $\kappa = 1$ and $B_k^T = \begin{bmatrix}1 & 1 & 0\end{bmatrix}$.

\begin{figure}
\centering

\begin{subfigure}{0.128\textwidth}
\includegraphics[width=\textwidth]{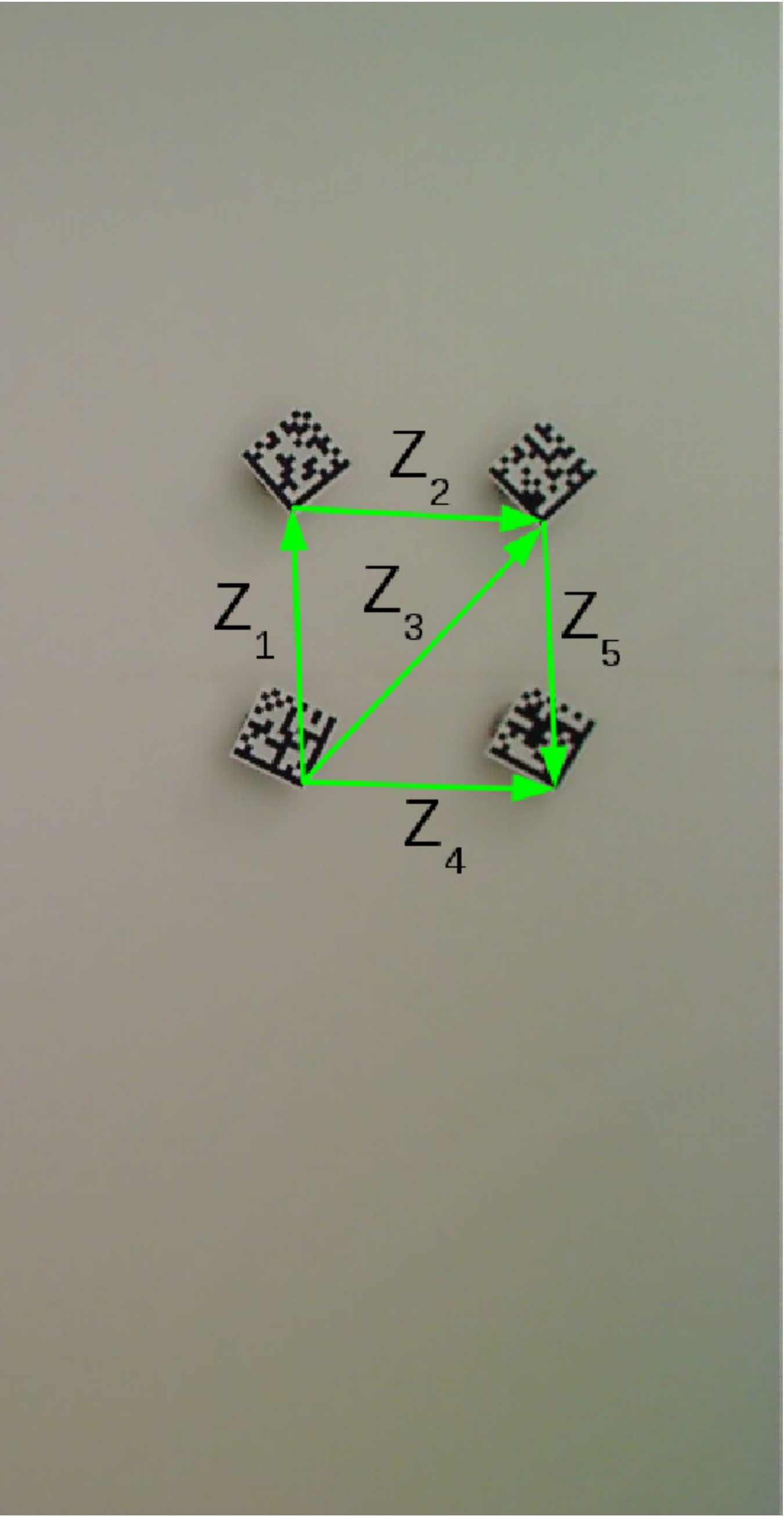}
\caption{}
\end{subfigure}
\begin{subfigure}{0.128\textwidth}
\includegraphics[width=\textwidth]{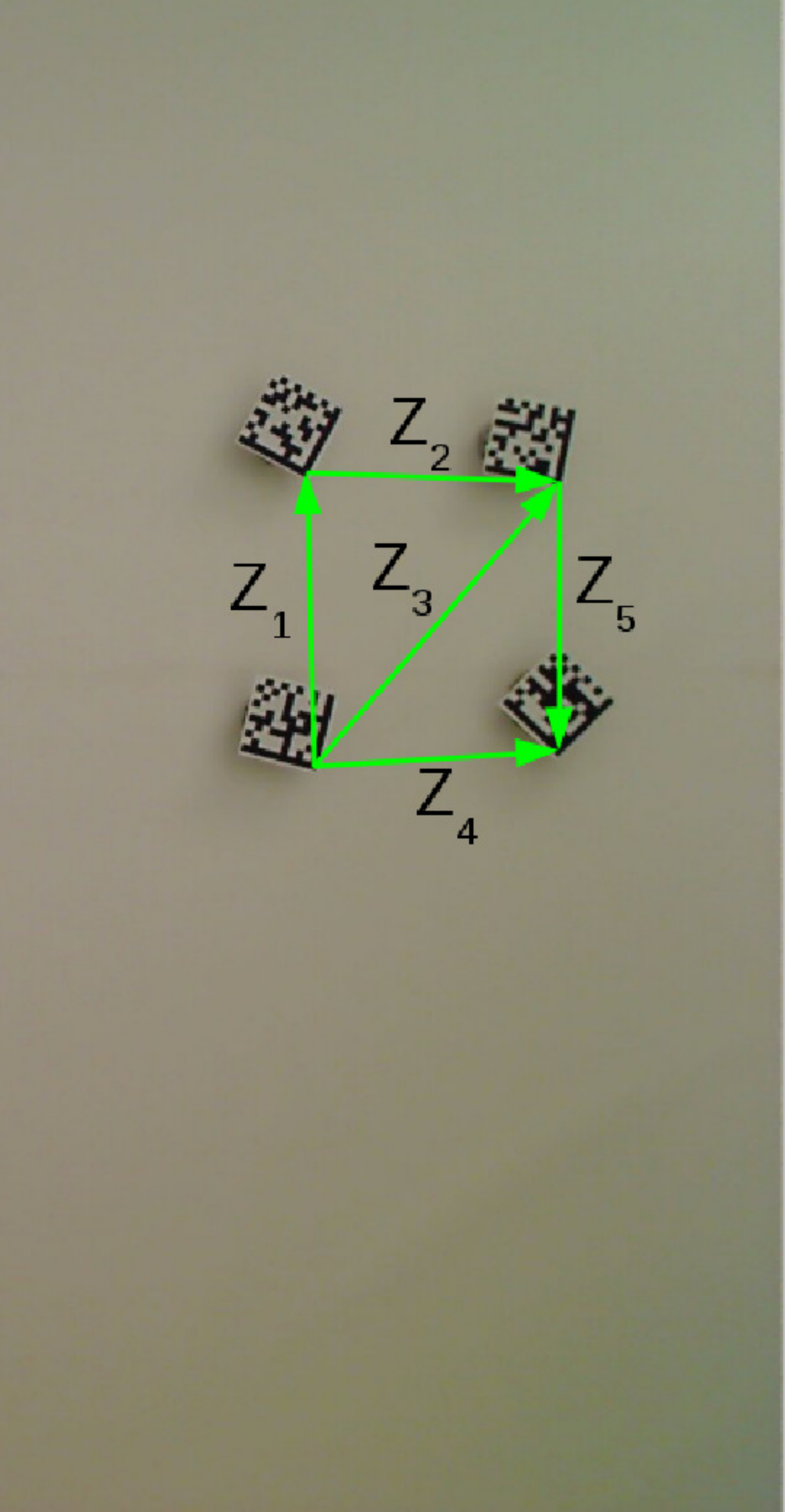}
\caption{}
\end{subfigure}
\begin{subfigure}{0.218\textwidth}
\includegraphics[width=\textwidth]{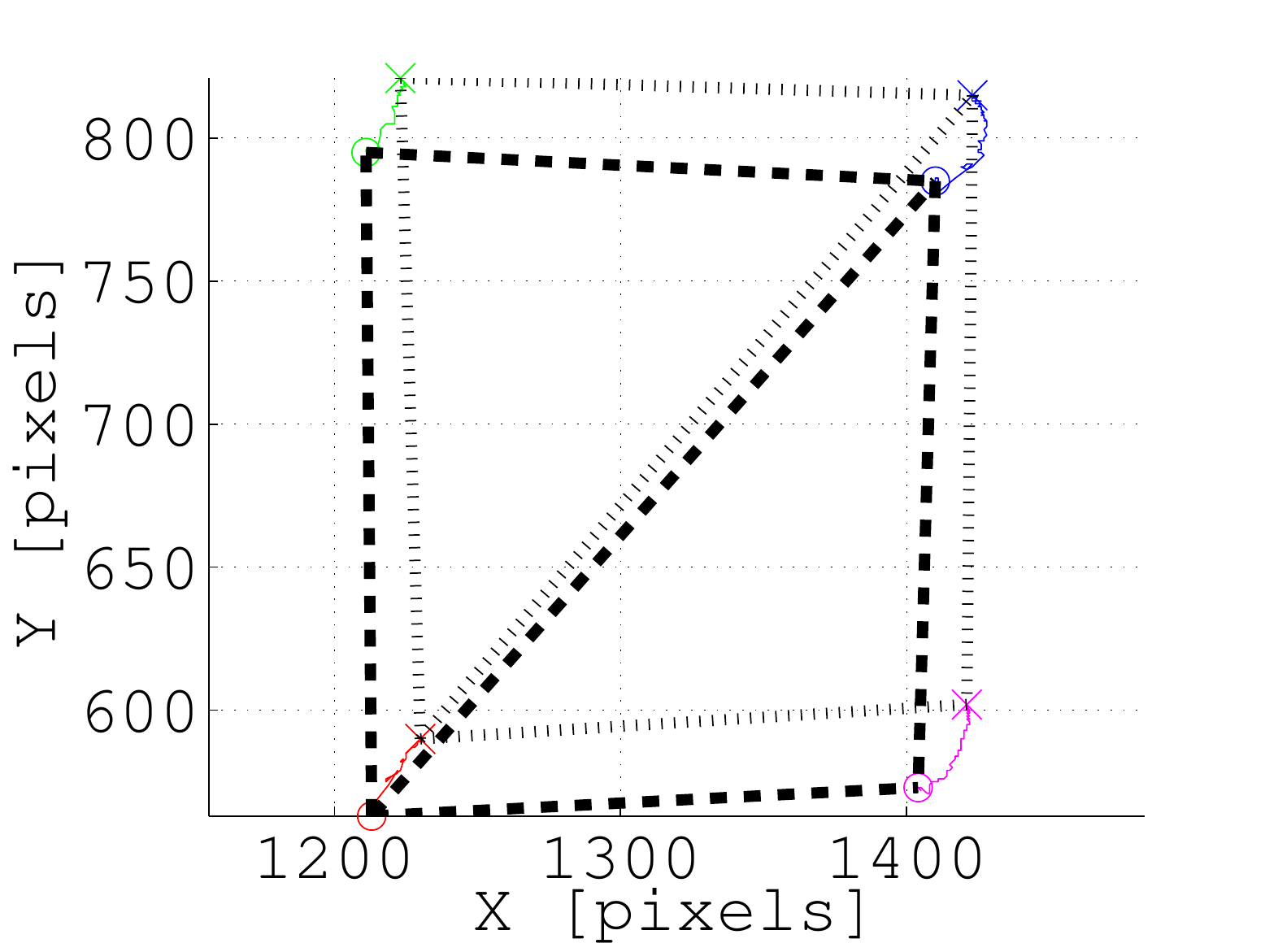}
\caption{}
\end{subfigure}
\begin{subfigure}{0.241\textwidth}
\includegraphics[width=\textwidth]{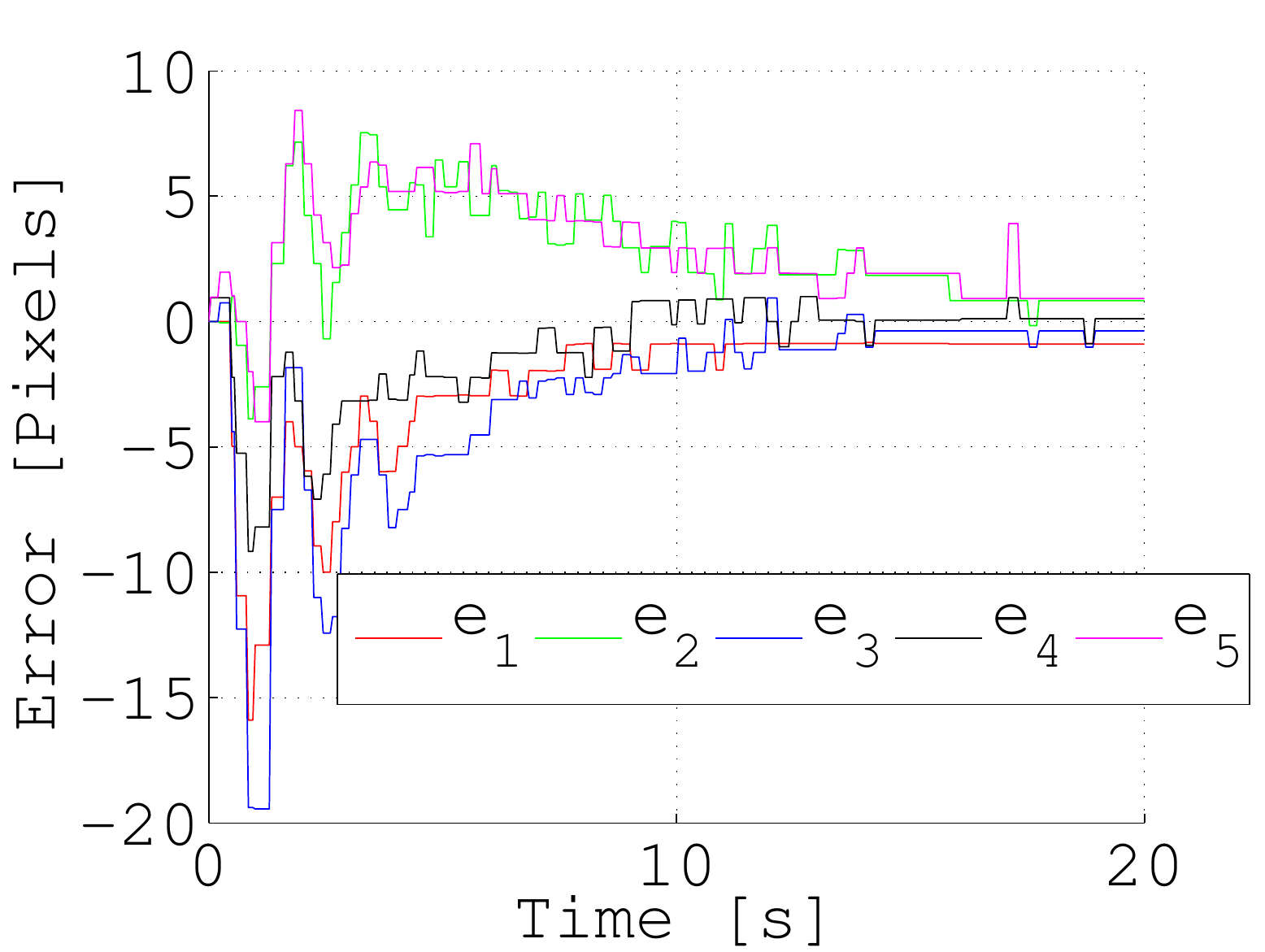}
\caption{}
\end{subfigure}
\begin{subfigure}{0.241\textwidth}
\includegraphics[width=\textwidth]{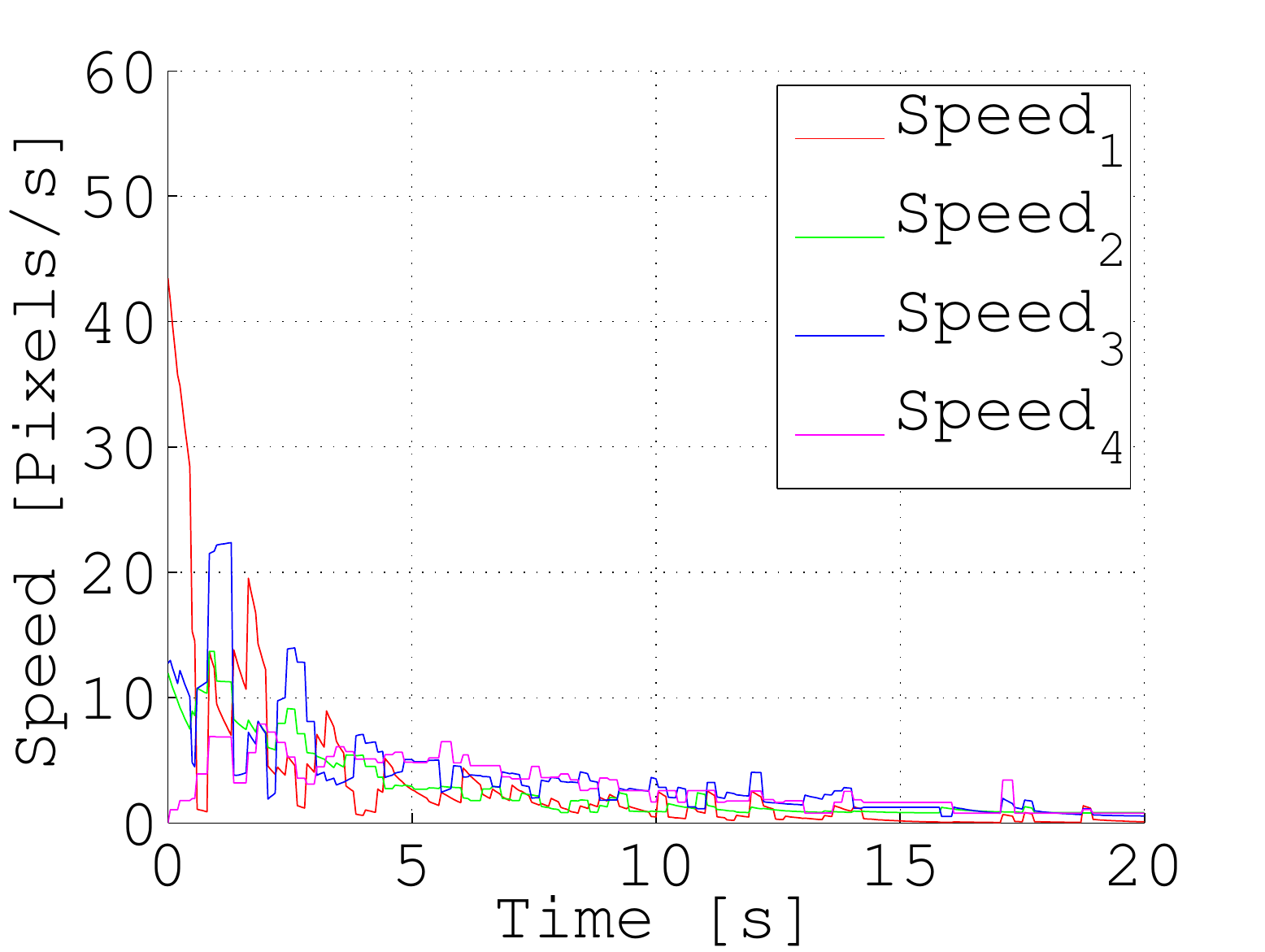}
\caption{}
\end{subfigure}
\begin{subfigure}{0.48\textwidth}
\includegraphics[width=\textwidth]{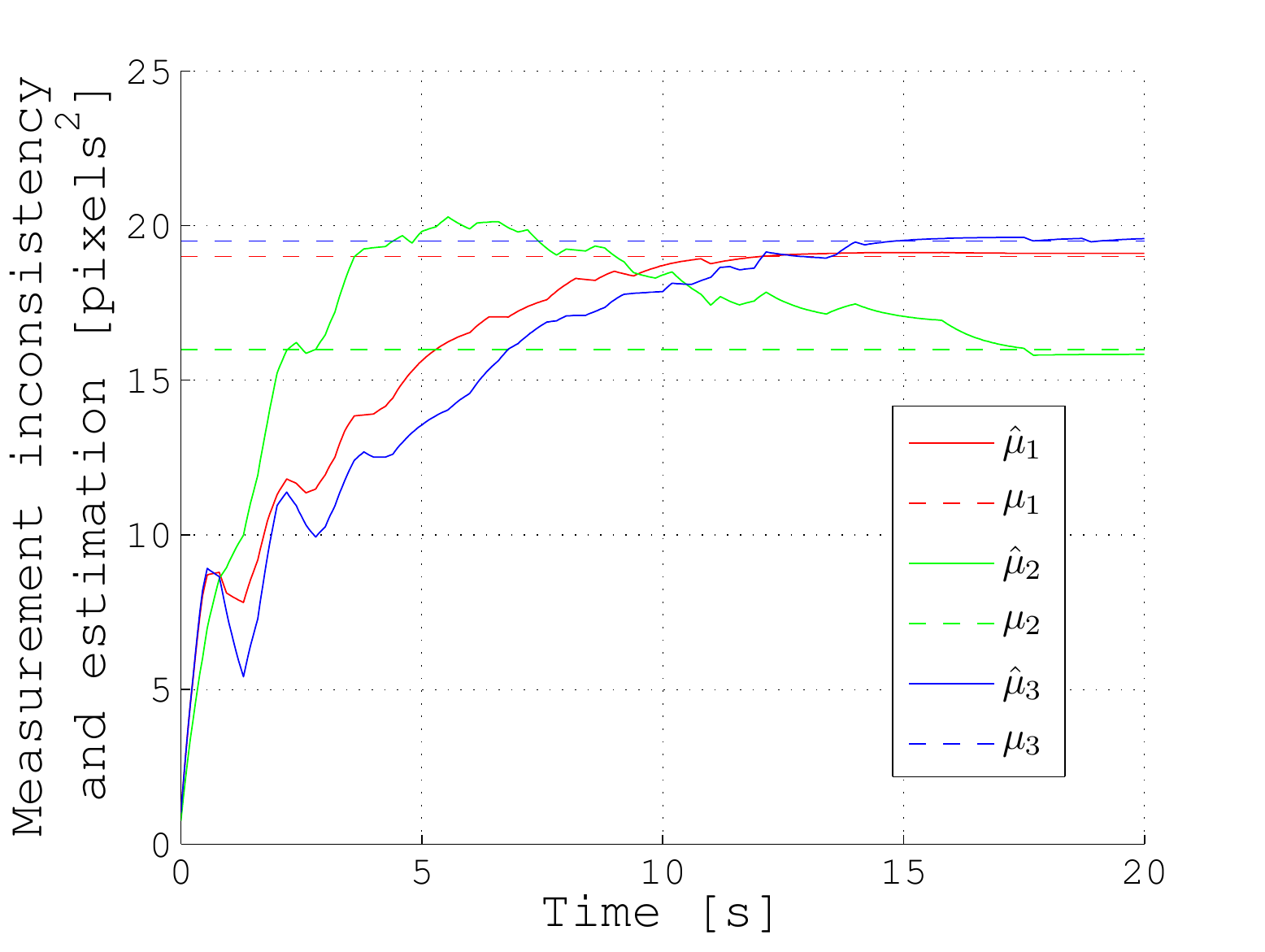}
\caption{}
\end{subfigure}

\caption{Experimental results of a 2D formation with inconsistent measurements using our proposed estimator-based gradient control. The initial configuration in (a) and the final configuration after 20 seconds in (b) correspond to the trajectory plot in (c) using dashed-lines and dotted-lines respectively; (d) the plot of errors $e_k$, $k=1,\dots, 5$; (e) the plots of the robots' speeds which show that the four agents eventually stop; (f) the plot of the estimated measurement inconsistencies for the first three edges (shown in solid-line) which asymptotically converge to the actual ones (shown in dashed-line).}
\label{fig: muhatXexp}
\end{figure}

In Figure \ref{fig: muhatXexp} we show the experimental result of the formation under the estimator-based gradient control. It is clear that the robots do not exhibit any undesirable motion induced by measurement inconsistency. The errors and the robots' speeds  converge to zero as soon as the inconsistency is effectively estimated by the estimating agents. In experiments, the errors and the speeds do not converge to zero precisely since once the discrepancies are approaching being effectively estimated, the control inputs become small and can be  dominated by  friction forces.

\subsection{Formation simulations in $\R^3$}
\begin{figure}[h!]
\centering
\begin{subfigure}{0.48\textwidth}
\includegraphics[width=\textwidth]{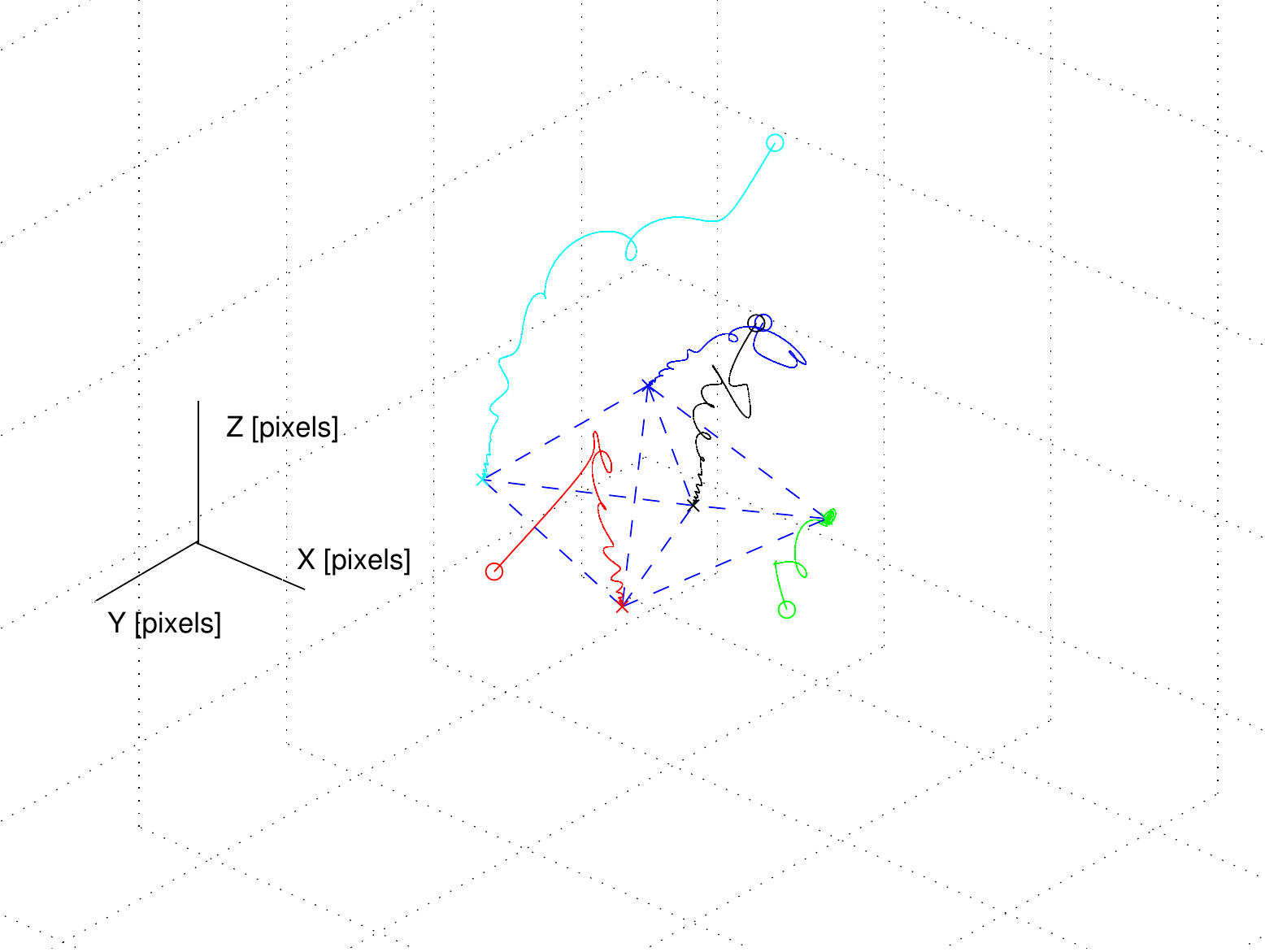}
\caption{}
\end{subfigure}
\begin{subfigure}{0.241\textwidth}
\includegraphics[width=\textwidth]{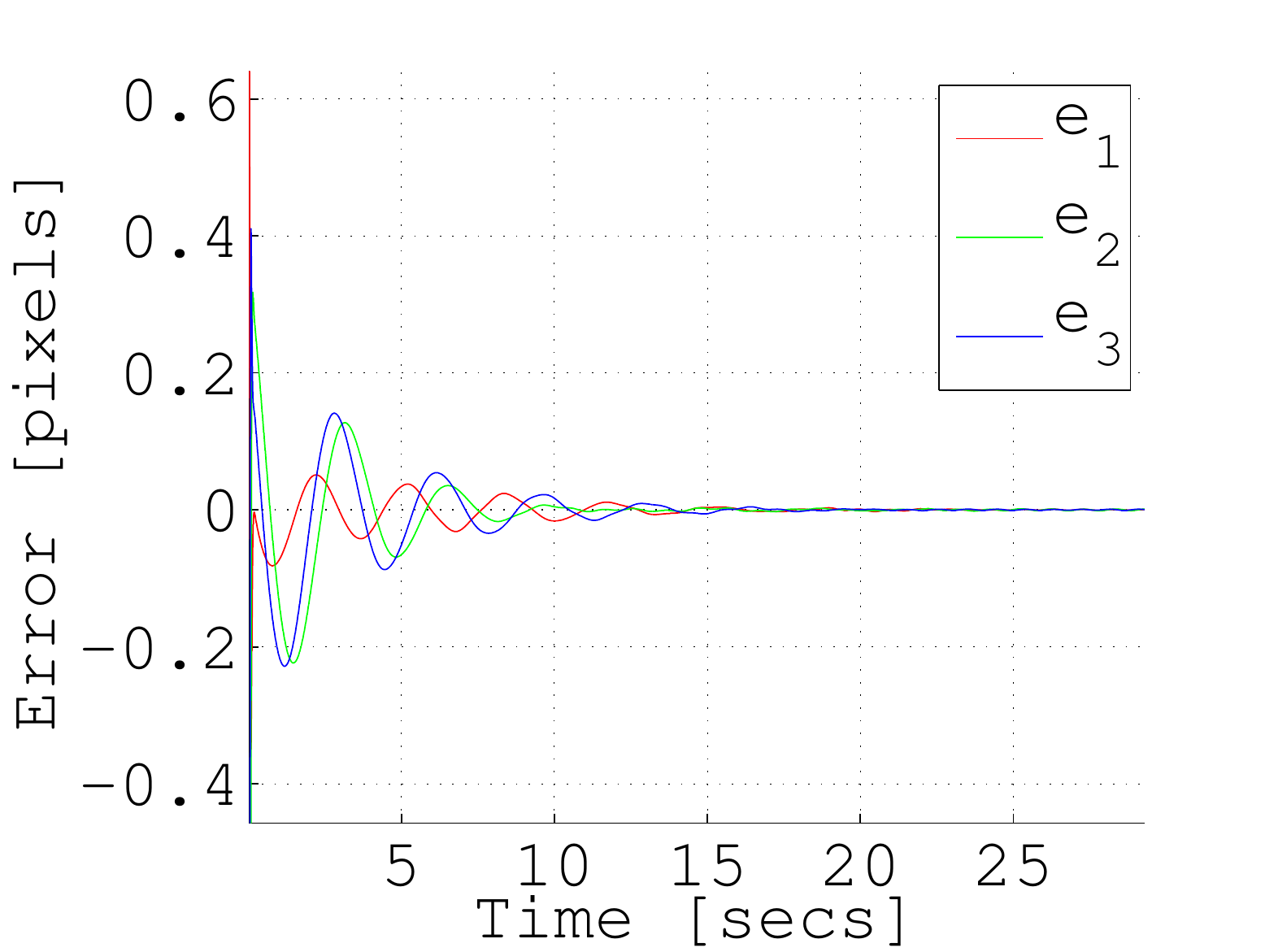}
\caption{}
\end{subfigure}
\begin{subfigure}{0.241\textwidth}
\includegraphics[width=\textwidth]{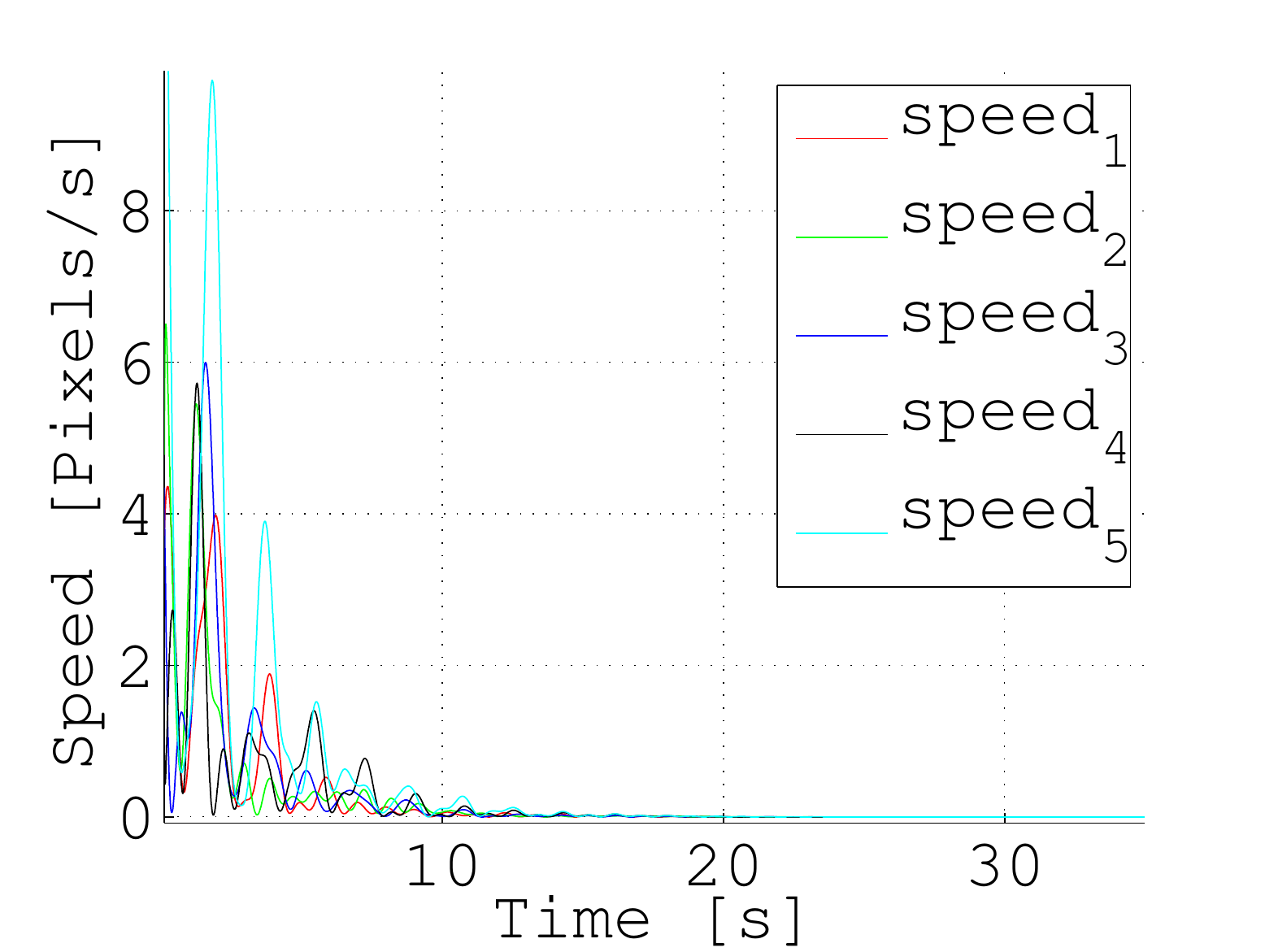}
\caption{}
\end{subfigure}
\begin{subfigure}{0.48\textwidth}
\includegraphics[width=\textwidth]{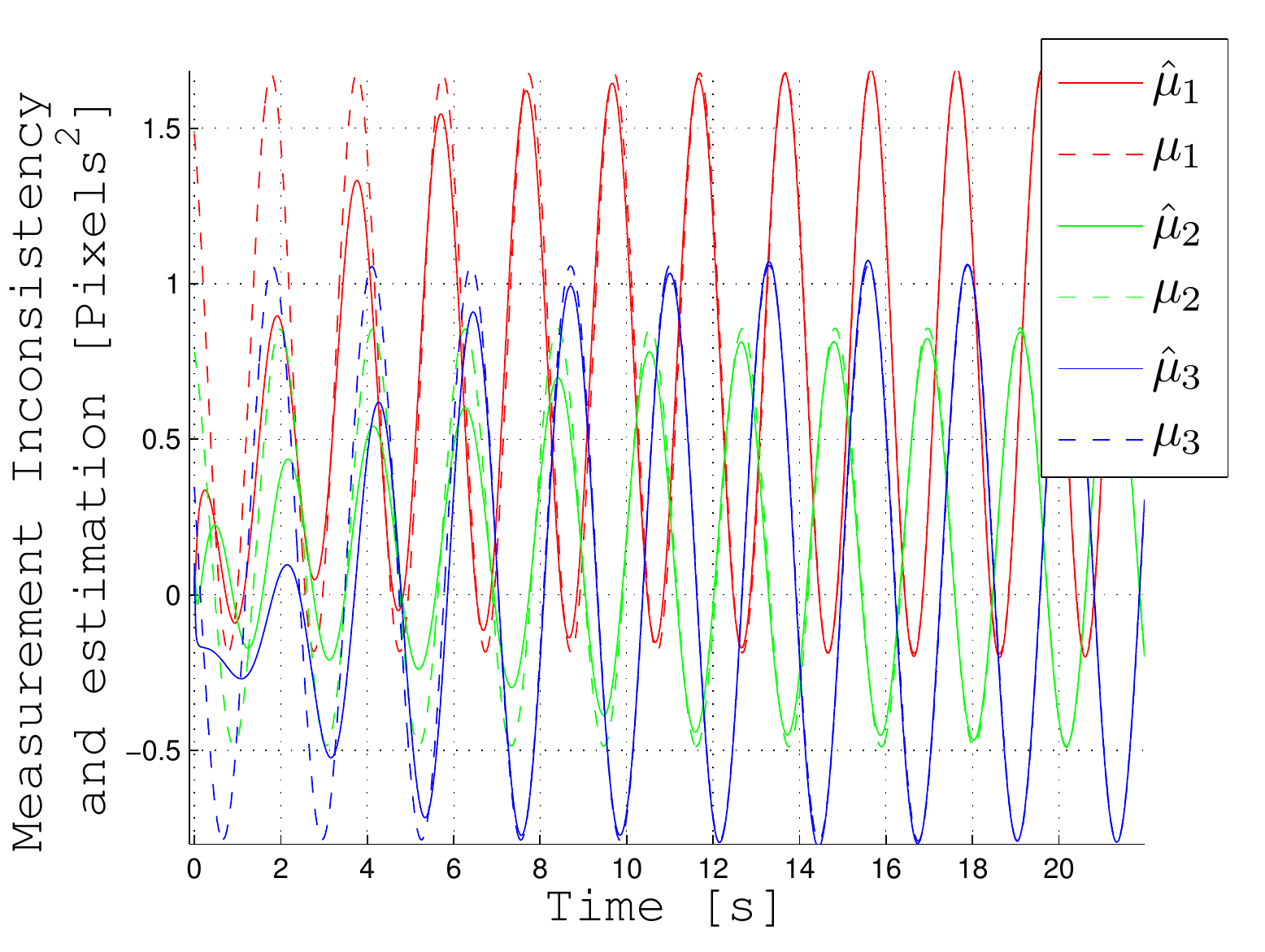}
\caption{}
\end{subfigure}
\caption{Simulation results of a 3D formation with inconsistent measurements using our proposed estimator-based gradient control where the inconsistencies are biased sinusoidal signals; (a) the plot of the trajectories where the initial positions are shown in circles and the final positions are shown in crosses; (b) the error $e_k$, $k=1, 2, 3$; (c) the plot of the agents' speeds which shows that the five agents eventually stop; (d) the plot of the estimated measurements inconsistencies for the first three edges (shown in solid-line) which asymptotically converge to the actual ones (shown in dashed-line).}
\label{fig: tX1}
\end{figure}

In this numerical setup, we consider a formation of five agents in $\R^3$. The measurement inconsistency takes the form of the superposition of a constant random offset and a sine wave with a known frequency. Each inconsistency $\mu_k$ has different frequencies and offsets. We implement control (\ref{eq: krick with muhat})-(\ref{eq: krick with muhat3}) and choose the estimating agents according to Proposition 5.5. The five agents are prescribed to maintain two regular tetrahedrons sharing the same base where all the edge lengths are $d = 5$. The five agents are placed randomly within a volume of 50 cubic units. We choose $\kappa=1$,  $B_k^T = \begin{bmatrix}1 & 1 & 0\end{bmatrix}$, and the estimators $\xi_k$ are initialized to be zero. In Figure \ref{fig: tX1} it is clear that the agents' velocities converge to zero and the formation shape converges to the desired one. Moreover, $\hat\mu(t)$ converges to the periodic inconsistency $\mu(t)$.

\section{Conclusions}
In this paper we have presented an estimator-based gradient control for stabilizing rigid formations using the internal model principle. We have effectively dealt with measurement inconsistency in the form of the combination of periodic signals with known frequencies but unknown amplitudes, phases and offsets. The proposed distributed control removes the surprising undesirable  steady-state movement reported by some recent papers. To carry out a key step of choosing estimating agents in our proposed approach, we have discussed a systematic way to guarantee the performance of our controller for classes of infinitesimally minimally rigid formations in $\R^2$ and $\R^3$. Experimental results for four mobile robots have been performed for formations in $\R^2$ and numerical simulations have been done for formations in $\R^3$. We are currently working on extending our control design to more detailed robot models, such as higher-order integrator and unicycle models. We are also interested in testing the performances of the controllers using outdoor robotic setups.


%





\ifCLASSOPTIONcaptionsoff
  \newpage
\fi



\bibliographystyle{IEEEtran}
\bibliography{tor,references_from_TOR,ref_ming}
%



%

\begin{IEEEbiography}[{\includegraphics[width=1in,height=1.25in,clip,keepaspectratio]{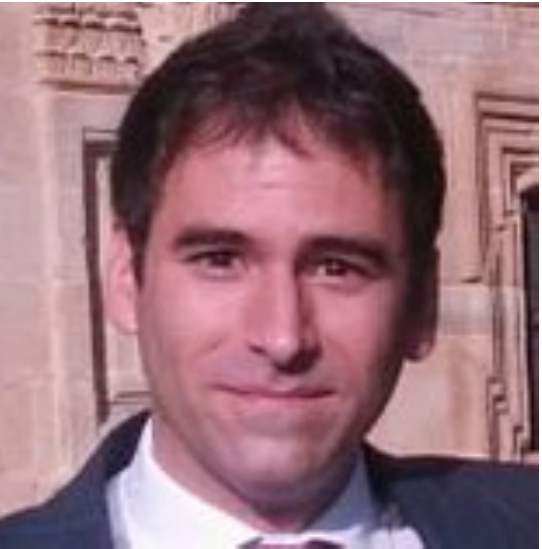}}]{Hector G. de Marina}
received the M.Sc. degree in electronics engineering from Complutense University of Madrid, Madrid, Spain in 2008 and the M.Sc. degree in control engineering from the University of Alcala de Henares, Alcala de Henares, Spain in 2011. He is currently a PhD student in the Faculty of Mathematics and Natural Sciences, University of Groningen, Groningen, The Netherlands. His research interests include formation control and navigation for autonomous robots.
\end{IEEEbiography}

\begin{IEEEbiography}[{\includegraphics[width=1in,height=1.25in,clip,keepaspectratio]{./images/ming}}]{Ming Cao}
received the PhD degree in electrical engineering from Yale University, New Haven, CT, USA in 2007. He is an associate professor with tenure responsible for the research direction of network analysis and control at the University of Groningen, the Netherlands. His main research interest is in autonomous agents and multi-agent systems, mobile sensor networks and complex networks.  He is an associate editor for \emph{Systems and Control Letters}, and for the Conference Editorial Board of the IEEE Control Systems Society. He is also a member of the IFAC Technical Committee on Networked Systems.
\end{IEEEbiography}

\begin{IEEEbiography}[{\includegraphics[width=1in,height=1.25in,clip,keepaspectratio]{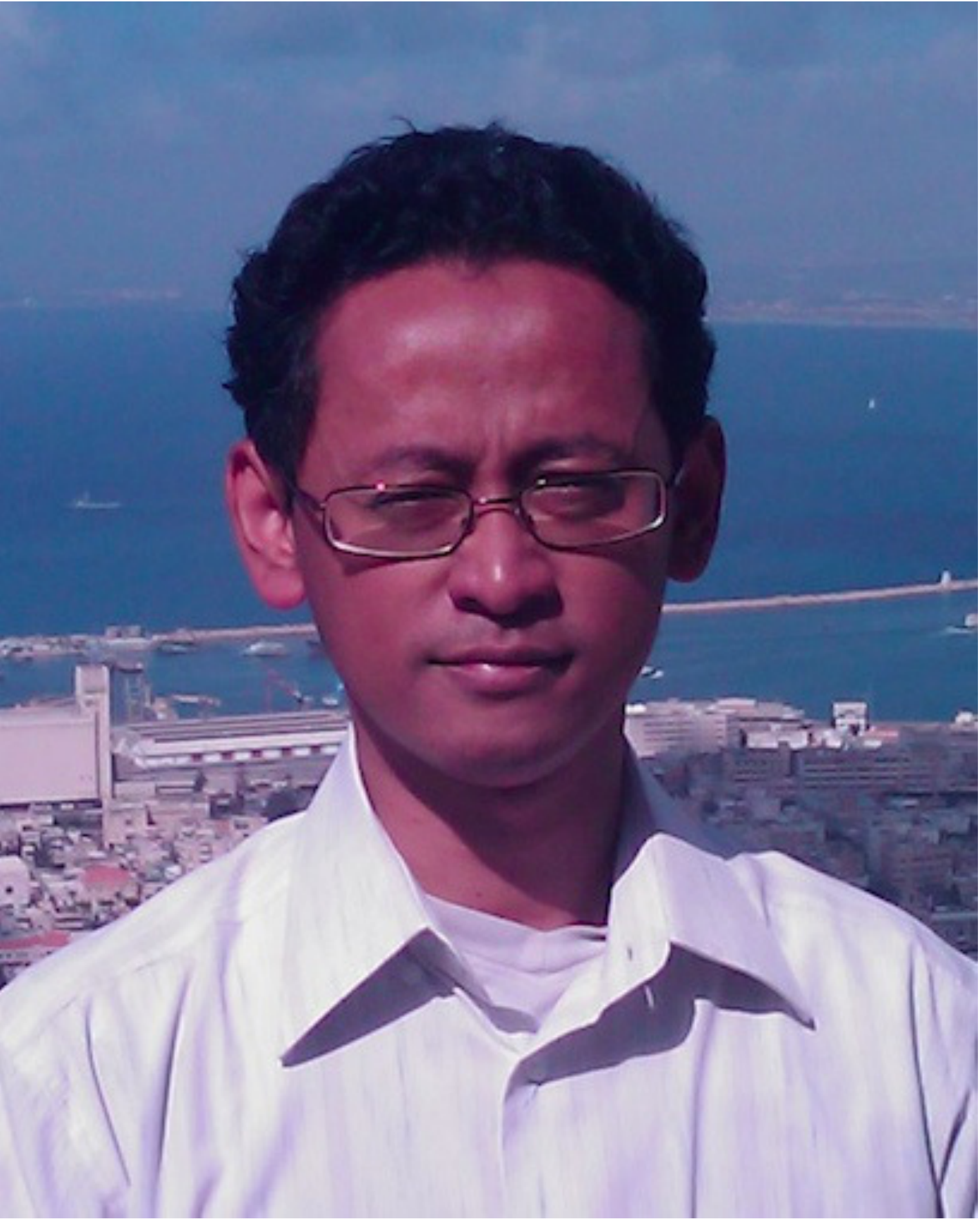}}]{Bayu Jayawardhana}
(SM’13) received the B.Sc. degree in electrical and electronics engineering from the Institut Teknologi Bandung, Bandung, Indonesia, in 2000, the M.Eng. degree in electrical and electronics engineering from the Nanyang Technological University, Singapore, in 2003, and the Ph.D. degree in electrical and electronics engineering from Imperial College London, London, U.K., in 2006. Currently, he is an associate professor in the Faculty of Mathematics and Natural Sciences, University of Groningen, Groningen, The Netherlands. He was with Bath University, Bath, U.K., and with Manchester Interdisciplinary Biocentre, University of Manchester, Manchester, U.K. His research interests are on the analysis of nonlinear systems, systems with hysteresis, mechatronics, systems and synthetic biology. 
Prof. Jayawardhana is a Subject Editor of the International Journal of Robust and Nonlinear Control, an associate editor of the European Journal of Control, and a member of the Conference Editorial Board of the IEEE Control Systems Society.
\end{IEEEbiography}







\end{document}